\documentclass[conference]{IEEEtran}
\IEEEoverridecommandlockouts
\usepackage{cite}
\usepackage{amsmath,amssymb,amsfonts}
\usepackage{algorithmic}
\usepackage{graphicx}
\usepackage{textcomp}
\usepackage{xcolor}
\usepackage{cuted}
\usepackage{lipsum}
\usepackage{mathtools}
\usepackage{caption}
\usepackage{float}

\begin{document}

	\title{Inverse Quantum Fourier Transform Inspired Algorithm for Unsupervised Image Segmentation
	}
	
	\author{\IEEEauthorblockN{$^{\dag}$Taoreed Akinola, $^{\dag}$Xiangfang Li, $^{\dag}$Richard Wilkins, $^{\dag}$Pamela Obiomon,  $^{\dag\ddag}$Lijun Qian}
		\IEEEauthorblockA{$^{\dag}$Department of Electrical and Computer Engineering, Prairie View A\&M University \\
			Prairie View, Texas 77446, USA \\
			Email: takinola2@pvamu.edu,  xili@pvamu.edu, rtwilkins@pvamu.edu, phobiomon@pvamu.edu, liqian@pvamu.edu}
			$^{\ddag}$ corresponding author
	}
	\maketitle

	\begin{abstract}
	Image segmentation is a very popular and important task in computer vision. In this paper, inverse quantum Fourier transform (IQFT) for image segmentation has been explored and a novel IQFT-inspired algorithm is proposed and implemented by leveraging the underlying mathematical structure of the IQFT. 
	 Specifically, the proposed method takes advantage of the phase information of the pixels in the image by encoding the pixels' intensity into qubit relative phases and applying IQFT to classify the pixels into different segments automatically and efficiently. 
 To the best of our knowledge, this is the first attempt of using IQFT for unsupervised image segmentation. The proposed method has low computational cost comparing to the deep learning based methods and more importantly it does \emph{not} require training, thus make it suitable for real-time applications.
The performance of the proposed method is compared with K-means and Otsu-thresholding. 
The proposed method outperform both of them on the PASCAL VOC 2012 segmentation benchmark and  the xVIEW2 challenge dataset by as much as $50\%$ in terms of mean Intersection-Over-Union (mIOU). 
	\end{abstract}

\begin{IEEEkeywords}
	\textit{Inverse Quantum Fourier Transform, Computer Vision, Image Segmentation} 
\end{IEEEkeywords}

\section{Introduction}\label{sec:introduction}
Image segmentation is defined as the separation of an image dataset into non-intersecting, homogeneous subsections in terms of certain properties such as color intensities and textures~\cite{Color_image_segmentation_cheng2001}. In image processing and computer vision, segmentation is generally applied  to separate the regions of interest  from the other part of the image (called the background)  for further analysis \cite{A_comparative_evaluation_for_liver_segmentation_from_spir_images_Goceri2013}. Segmentation is a crucial and difficult problem in many fields such as digital image processing,  object or pattern recognition, enhanced object features extraction and artificial intelligence~\cite{Image_segmentation_evaluation_wang2020,Segmentation_Techniques_for_Complex_Image_pandey2020}. It is a very important tool for  solving image semantic problems which involves predicting the categories of the pixels contained in an image \cite{Image_semantic_segmentation_method_based_on_improved_ERFNet_model_ye2022}. The applications of image segmentation span many different fields, such as  in medical diagnostic imaging applications for automatic detection of anomalies, treatment monitoring and disease diagnosis \cite{ High_resolution_encoder_decoder_networks_for_low_contrast_medical_image_segmentation_Zhou2019,  Biomedical_information_technology_image_based_computer_aided_diagnosis_systems_goceri2018}, in image-based remote sensing technology~\cite{Hybrid_remote_sensing_image_segmentation_considering_intrasegment_homogeneity_and_intersegment_heterogeneity_wang2019}, and in additive manufacturing processes for in-situ monitoring and detection of shifts in product quality during production \cite{Image_Decomposition_Accelerates_Dynamic_Network_Modeling_for_in_situ_Monitoring_of_Bio_mimic_Wing_Printing_Processes_aworunse2019}. 

In the classical computing domain, Fourier transform is one of the popular techniques that have been researched and developed extensively for image segmentation \cite{Texture_identification_and_image_segmentation_via_Fourier_transform_Zou2001, Joint_segmentation_and_nonlinear_registration_using_fft_and_total_variation_atta2018, Coffee_plantation_area_recognition_in_satellite_images_using_Fourier_transformtsai2017}.
For instance, the auto-registration property of the magnitude spectra has been exploited for texture identification~\cite{Texture_identification_and_image_segmentation_via_Fourier_transform_Zou2001}.
Similarly, structural texture in the satellite images have been extracted using Fourier transform for agriculture applications~\cite{Coffee_plantation_area_recognition_in_satellite_images_using_Fourier_transformtsai2017}.
Segmenting foreground to track spontaneous changes in the shape of objects embedded in a template image using total variation and fast Fourier transform has been studied in~\cite{Joint_segmentation_and_nonlinear_registration_using_fft_and_total_variation_atta2018}.
On the other hand, several image segmentation algorithms have emerged in the quantum computing domain. For instance, the authors of~\cite{The_dual_threshold_quantum_image_segmentation_algorithm_and_its_simulation_yuan2020}  provided a quantum circuit for gray image encoding and a proof-of-concept quantum circuit of dual-threshold segmentation algorithm, and simulated it on an $8\times8$ image.
However, inverse quantum Fourier transform (IQFT) for image segmentation has not been studied.

In this work, we seek to leverage the unique properties of IQFT and provide an IQFT-inspired approach for unsupervised image segmentation. 
Specifically, the proposed method takes advantage of the phase information of the pixels in the image by encoding the pixels' intensity into qubit relative phases and applying IQFT to classify the pixels into different segments automatically and efficiently. 
The contributions of this paper are: 
\begin{enumerate}
	\item {An IQFT-inspired algorithm is proposed for unsupervised image segmentation. \emph{To the best of our knowledge, this is the first unsupervised image segmentation algorithm using IQFT.}}
	\item {Comparing to many deep learning based image segmentation methods, the proposed approach has much less computational cost. More importantly, the proposed method does \emph{not} require training, thus make it suitable for real-time applications.}
	\item {We demonstrate the potential of IQFT in solving image segmentation problem, and  compare its effectiveness with two popular unsupervised segmentation techniques: Otsu thresholding~\cite{A_threshold_selection_method_from_graylevel_histograms_otsu1979}, and K-means Clustering~\cite{Classification_and_analysis_of_multivariate_observations_macqueen1967}. The proposed method outperform both of them on
	the PASCAL VOC 2012 segmentation benchmark~\cite{Unsupervised_learning_of_image_segmentation_based_on_differentiable_feature_clustering_kim2020}, and  the xVIEW2 challenge dataset~\cite{xbd_A_dataset_for_assessing_building_damage_from_satellite_imagery_gupta2019} by as much as $50\%$ in terms of mIOU. }
	\end{enumerate}
The remainder of this paper is structured as follows: Related works are reviewed in Section~\ref{sec:relatedworks}. Section~\ref{sec:QIFT} introduces inverse quantum Fourier transform. Section~\ref{sec:QuantumInverseFourierTransformRGB} provides the details of the proposed approach. The experimental results are given in Section~\ref{sec:results} and observations and insights from the results are discussed. Section~\ref{sec:concl} concludes the paper. 

\section{Related Works}\label{sec:relatedworks}
\subsection{Image Segmentation in Classical Computing}\label{subsec:ImageSegmentationClassicalComputing}
As a key process in image processing and analysis, image segmentation has been well studied in the classical computing domain. Many methods for solving image segmentation problems have emerged over the years. These methods vary widely depending on the specific application since a single method is not sufficient for different images with varying characteristics in terms of sharpness, texture, noise presence, and the degree of overlapping objects~\cite{Segmentation_Techniques_for_Complex_Image_pandey2020}. Traditional techniques used for image segmentation are categorized as thresholding-based technique~\cite{A_comparative_performance_study_of_several_global_thresholding_techniques_for_segmentation_lee1990}, region-based technique~\cite{Region_based_segmentation_versus_edge_detection_kaganami2009}, edge-based technique~\cite{Edge_region_based_segmentation_of_range_images_wani1994}, clustering-based technique~\cite{Survey_on_clustering_based_image_segmentation_techniques_zou2016}, and watershed technique~\cite{Image_segmentation_based_on_watershed_and_edge_detection_techniques_salman2006}. These methods vary widely depending on the specific application since they all have their limitations based on the underlying principles. For instance, employing an unsupervised method like the K-means for image segmentation has a major drawback which is the requirement for the optimal number of clusters to be specified before the algorithm is applied~\cite{Selection_of_K_in_K_means_clustering_pham2005}. The segmentation error decreases as the number of clusters increases and there is no theoretical means of obtaining the optimal number of clusters to be used. Similarly, Otsu's thresholding technique does not consider the spatial information of image, and this makes it
sensitive to the unevenness and noise in a grayscale image~\cite{ A_review_on_image_segmentation_techniques_pal1993}. 

Although  the advent of deep learning has brought about new classes of image segmentation techniques that have become widely available~\cite{A_review_on_deep_learning_techniques_applied_to_semantic_segmentation_garcia2017}, usually they have very high computational complexity. Furthermore, most of them are supervised methods that require training and probably re-training when applied to a new dataset. On the contrary, the proposed method has low computational cost comparing to the deep learning based methods and more importantly it does \emph{not} require training, thus make it suitable for real-time applications.
In our analysis, we will compare the performance of our proposed technique with two popular unsupervised segmentation techniques that do not require training: Otsu thresholding~\cite{A_threshold_selection_method_from_graylevel_histograms_otsu1979}, and K-means Clustering~\cite{Classification_and_analysis_of_multivariate_observations_macqueen1967}.

\subsection{Image Segmentation in Quantum Computing}\label{subsec:ImageSegmentationQuantumComputing}
In recent years, several image segmentation algorithms have emerged in the quantum domain to exploit the properties of quantum computing to improve the performance of classical techniques and, subsequently, their applications. Some common algorithms are thresholding segmentation~\cite{Design_of_threshold_segmentation_method_for_quantum_image_Li2020}  and quantum search algorithm~\cite{Image_storage_retrieval_compression_and_segmentation_in_a_quantum_system_li2013}. Most of these proposed methods involve implementation of some oracle operators which are either fully theoretical or hard to simulate due to the number of qubits required. 

Most of the well-known quantum algorithms such as quantum phase estimation, Grover's algorithm and Shor's algorithm utilize QFT (or IQFT) as part of their major subroutines~\cite{Quantum_Computation_Quantum_Information_Nielsen_Chuang_2011}. Similarly, in applications such as quantum image processing, error-correction, and encryption~\cite{Mixed_state_entanglement_and_quantum_error_correction_bennett1996, System_and_method_for_key_distribution_using_quantum_cryptography_townsend1997}, QFT (or IQFT) is embedded. 
Consequently, QFT is undoubtedly a valuable transformation in the quantum domain, and itas applications are not yet fully studied~\cite{Implementation_and_Analysis_of_QFT_in_Image_Processing_al}. 
 
\section{Inverse Quantum  Fourier Transform }\label{sec:QIFT}
QFT transforms a computational basis state $\vert x \rangle$ into a superposition of all the computational basis states with the inclusion of relative phases. This is defined mathematically in equation~(\ref{eq:QFT1})~\cite{Quantum_Computation_Quantum_Information_Nielsen_Chuang_2011}, where $ N =2^n $, $ n$ is the number of  qubits; $\omega=e^{i\frac{2\pi}{N}}$, the $\ nth$ root of  unity. In tensor product form, equation~(\ref{eq:QFT1}) becomes equation~(\ref{eq:QFT2})~\cite{Quantum_Computation_Quantum_Information_Nielsen_Chuang_2011}. Considering a three qubit system, for example, $n = 3$, equation~(\ref{eq:QFT2}) can be expanded to equation~(\ref{eq:QFT3}).
\begin{equation}
	\begin{aligned}
		Q F T(|x\rangle)=\frac{1}{\sqrt{N}} \sum_{k=0}^{N-1} \omega^{x k}|k\rangle\hfill
		\label{eq:QFT1}
	\end{aligned}
\end{equation}
\begin{equation}
	\begin{aligned}
		\centering
		Q F T(|x\rangle _n)=\frac{1}{\sqrt{N}} \otimes_{k=1}^{n}\left(|0\rangle+e^{i \frac{2 \pi x}{2^k}}|1\rangle\right)
		\label{eq:QFT2}
	\end{aligned}
\end{equation}
\begin{align}
	Q F T(|x\rangle _3)=\frac{1}{\sqrt{8}} \left(|0\rangle+e^{i \frac{2 \pi x}{2}}|1\rangle\right)\otimes \left(|0\rangle+e^{i \frac{2 \pi x}{4}}|1\rangle\right)  \nonumber\\ \otimes \left(|0\rangle+e^{i \frac{2 \pi x}{8}}|1\rangle\right)
	\label{eq:QFT3}
\end{align}
For example, the QFT of a quantum state $\vert x \rangle=|100\rangle$  is determined from equation~(\ref{eq:QFT3})  by substituting $ x= 4$, the decimal equivalent of $100_2$. This operation is  shown in equation~(\ref{eq:QFT4}). The result suggests that a superposition of states, with some phase information, can be transformed into a single state representation by employing the inverse operation of the QFT.
\begin{IEEEeqnarray}{l}
	Q F T(|x=100\rangle _3) =\frac{1}{\sqrt{8}} (|000\rangle-|001\rangle+|010\rangle-|011\rangle\nonumber\\+|100\rangle-|101\rangle+|110\rangle-|111\rangle )
	\label{eq:QFT4}
\end{IEEEeqnarray}
The  inverse quantum Fourier transform (IQFT) performs the reverse operation of the QFT. It transforms from a phase representation into the computational basis. From  equation~(\ref{eq:QFT2}), the inverse operation, IQFT, can be written as equation~(\ref{eq:QFT5}), where $\vert k \rangle$ is the state of a quantum system with some phase information~\cite{Quantum_Computation_Quantum_Information_Nielsen_Chuang_2011}. Equation~(\ref{eq:QFT6}) presents a case of three qubits, $ n=3 $.
\begin{equation}
	\centering
	IQFT\bigg[\frac{1}{\sqrt{N}} \otimes_{k=1}^{n}\left(|0\rangle+e^{i \frac{2 \pi x}{2^k}}|1\rangle\right)\bigg]=|k\rangle _n\hfill
	\label{eq:QFT5}
\end{equation}
\begin{IEEEeqnarray}{l}
	IQFT\bigg[\frac{1}{\sqrt{8}} \left(|0\rangle+e^{i \frac{2 \pi x}{2}}|1\rangle\right)\otimes \left(|0\rangle+e^{i \frac{2 \pi x}{4}}|1\rangle\right)\nonumber \\ \hspace{7em}\otimes \left(|0\rangle+e^{i \frac{2 \pi x}{8}}|1\rangle\right)\bigg]=|k\rangle _3
	\label{eq:QFT6}
\end{IEEEeqnarray}
\section{Proposed IQFT-Inspired Algorithm for RGB Image Segmentation}\label{sec:QuantumInverseFourierTransformRGB}
\subsection{The underlying mathematics of the proposed method}
IQFT for image segmentation is founded on equation~(\ref{eq:QFT6}) because it utilizes three qubits that match the three channels of  RGB color space. Replacing the relative phases $\frac{2\pi x}{2}, \frac{2\pi x}{4}$ and $\frac{2\pi x}{8}$ with $\alpha, \beta$ and $\gamma$, respectively, results in equation~(\ref{eq:QFT7}). 
Expansion of this equation results in equation~(\ref{eq:QIFT10}). The terms  $IQF T|000\rangle$ is expanded as shown in equation~(\ref{eq:QIFT2}). The remaining terms are similarly expanded, and  substituted in equation~(\ref{eq:QIFT10}) to get equation~(\ref{eq:QIFT12}), where the coefficients P, Q, R, S, T, U, V, W are the probability amplitudes  defined in  equation~(\ref{eq:QIFT13}). This matrix formulation is the underlying equation of our proposed method.
\begin{strip}
\begin{equation}
	IQFT\bigg[\frac{1}{\sqrt{8}} \left(|0\rangle+e^{i \alpha }|1\rangle\right)\otimes \left(|0\rangle+e^{i \beta }|1\rangle\right) \otimes \left(|0\rangle+e^{i \gamma}|1\rangle\right)\bigg]=|k\rangle _3
	\label{eq:QFT7}
\end{equation}
	\begin{IEEEeqnarray}{l}
	IQF T \left(\frac{1}{\sqrt{8}}\left(|0\rangle+e^{i \alpha}|1\rangle\right) \otimes\left(|0\rangle+e^{i \beta}|1\rangle\right) \otimes\left(|0\rangle+e^{i \gamma}|1\rangle\right)\right)\nonumber\\ 
	=\frac{1}{\sqrt{8}} \times\left[IQF T|000\rangle+e^{i \gamma}IQF T|001\rangle+e^{i \beta} I Q F T|010\rangle+e^{i(\beta+\gamma)} I Q F T|011\rangle\right. \nonumber\\ 
	\hspace{7em}
	+e^{i \alpha}IQF T|100\rangle+e^{i(\alpha+\gamma)} IQF T|101\rangle+e^{i(\alpha+\beta)} IQF T|110\rangle
	\left.+e^{i(\alpha+\beta+\gamma)} IQF T|111\rangle\right]  
	\label{eq:QIFT10}
	\end{IEEEeqnarray}
	\begin{align}
		IQF T(|000\rangle) =\frac{1}{\sqrt{8}} \sum_{x=0}^{7} \omega^{-x(0)}|x\rangle =\frac{1}{\sqrt{8}}(|000\rangle+|001\rangle+|010\rangle+|011\rangle+|100\rangle+|101\rangle+|110\rangle+|111\rangle) 
		\label{eq:QIFT2}
	\end{align}
	\begin{IEEEeqnarray}{l}
		IQF T \left(\frac{1}{\sqrt{8}}\left(|0\rangle+e^{i \alpha}|1\rangle\right) \otimes\left(|0\rangle+e^{i \beta}|1\rangle\right) \otimes\left(|0\rangle+e^{i \gamma}|1\rangle\right)\right)	\hspace{17em}\nonumber\\ 
		\hspace{7em}= P|000\rangle+Q|001\rangle+R|010\rangle+S|011\rangle+T|100\rangle+U|101\rangle+V|110\rangle+W|111\rangle 
		\label{eq:QIFT12}
	\end{IEEEeqnarray}
	\begin{IEEEeqnarray}{l}
		\left[\begin{array}{c}
			P \\ Q \\ R \\ S \\ T\\ U \\ V \\W
		\end{array}\right] 
		\equiv~\frac{1}{8} \times \left[\begin{array}{cccccccc}
			1 & 1 & 1 & 1 & 1 & 1 & 1 & 1 \\
			1 & \omega^{-1} & \omega^{-2} & \omega^{-3} & \omega^{-4} & \omega^{-5} & \omega^{-6} & \omega^{-7} \\
			1 & \omega^{-2} & \omega^{-4} & \omega^{-6} & 1 & \omega^{-2} & \omega^{-4} & \omega^{-6} \\
			1 & \omega^{-3} & \omega^{-6} & \omega^{-1} & \omega^{-4} & \omega^{-7} & \omega^{-2} & \omega^{-5} \\
			1 & \omega^{-4} & 1 & \omega^{-4} & 1 & \omega^{-4} & 1& \omega^{-4}  \\
			1 & \omega^{-5} & \omega^{-2} & \omega^{-7} & \omega^{-4} & \omega^{-1} & \omega^{-6} & \omega^{-3} \\
			1 & \omega^{-6} & \omega^{-4} & \omega^{-2} & 1 & \omega^{-6} & \omega^{-4} & \omega^{-2} \\
			1 & \omega^{-7} & \omega^{-6} & \omega^{-5} & \omega^{-4} & \omega^{-3} & \omega^{-2} & \omega^{-1}
		\end{array}\right] \left[\begin{array}{c}
			1 \\
			e^{i \gamma} \\
			e^{i \beta} \\
			e^{i(\beta+\gamma)} \\
			e^{i \alpha} \\
			e^{i(\alpha+\gamma)} \\
			e^{i(\alpha+\beta)} \\
			e^{i(\alpha+\beta+\gamma)}
		\end{array}\right]
		\label{eq:QIFT13}
	\end{IEEEeqnarray}
\end{strip}
\begin{figure*}[h!]
	\centering
	\begin{minipage}{0.24\linewidth}
		\centering
		\includegraphics[width=\linewidth]{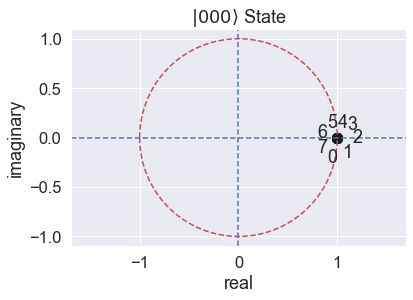} \\
		\includegraphics[width=\linewidth]{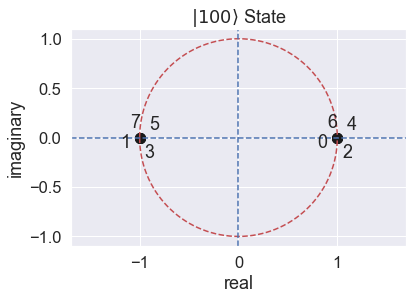} 
		\label{basis1}
	\end{minipage}
	\begin{minipage}{0.24\linewidth}
		\centering
		\includegraphics[width=\linewidth]{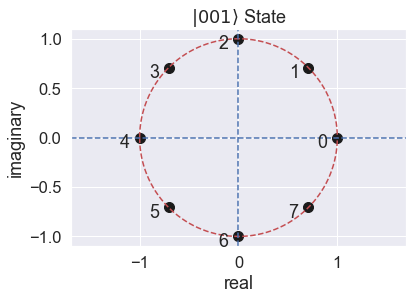} \\
		\includegraphics[width=\linewidth]{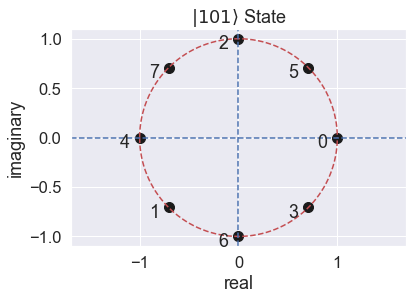} 
		\label{basis2}
	\end{minipage}
	\begin{minipage}{0.24\linewidth}
		\centering
		\includegraphics[width=\linewidth]{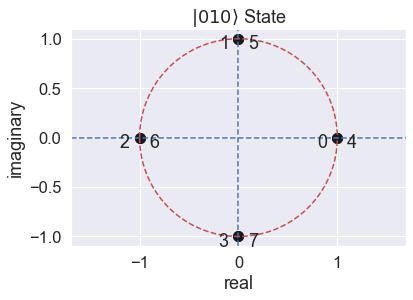} \\
		\includegraphics[width=\linewidth]{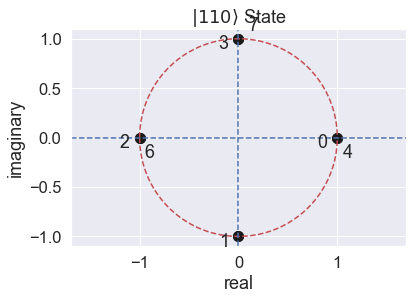} 
		\label{basis3}
	\end{minipage}
	\begin{minipage}{0.24\linewidth}
		\centering
		\includegraphics[width=\linewidth]{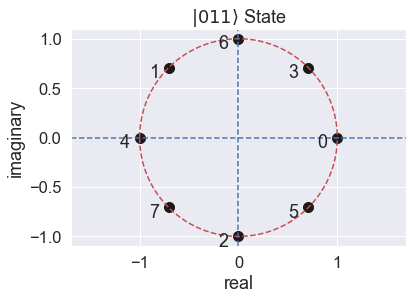} \\
		\includegraphics[width=\linewidth]{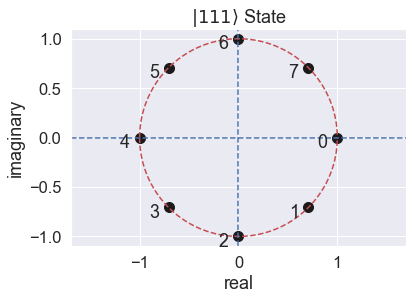} 
		\label{basis4}
	\end{minipage}
	\caption{Visualization of the eight state basis vectors}
	\label{fig:allbasis}
\end{figure*}

\begin{figure}[h!]
	\centering
	\includegraphics[width=0.7\linewidth]{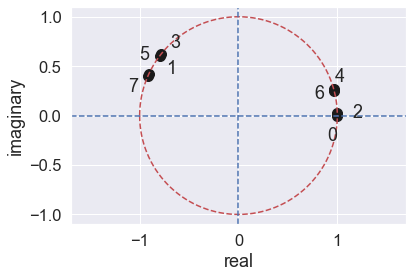} 
	\caption{Transformed input pattern on the unit circle for a random case of $\alpha =2.464$ , $\beta =0.025$, and $\gamma =0.246$. Some points are coincident. }
	\label{fig:ExamplesInputTransf}
\end{figure}

\begin{figure}[h!]
	\centering
	\includegraphics[width=0.7\linewidth]{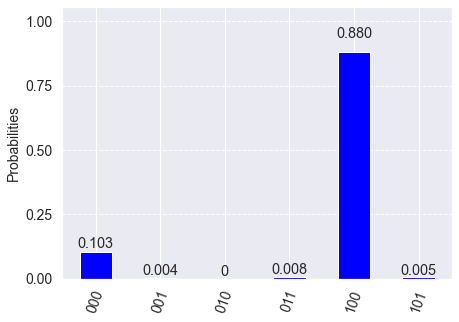} 
	\caption{Probability distribution for input pattern of a random case of $\alpha =2.464$ , $\beta =0.025$, and $\gamma =0.246$ }
	\label{fig:probDistrExamples}
\end{figure}

\subsection{Proposed IQFT-Inspired Algorithm}\label{subsec:IQFTInspiredAlgorithm}
The proposed algorithm is developed from equation~(\ref{eq:QIFT13}). Firstly, eight state basis vectors are given by the rows of the 8 by 8 matrix. These basis vectors are visualized as a set of points on a unit circle as shown in Figure~\ref{fig:allbasis}. Secondly, a 3-dimensional input vector, [$\alpha,\beta, \gamma$], is transformed into an eight-dimensional vector given by the column matrix on the right-hand side of equation~(\ref{eq:QIFT13}). This vector also can be represented by a set of eight points on a unit circle. A random example for which $\alpha =2.464$ , $\beta =0.025$, and $\gamma =0.246$ is shown in Figure~\ref{fig:ExamplesInputTransf}. By visual inspection, the input pattern has two obvious clusters similar to basis state vector $|100\rangle$. Lastly, the probability that the pattern generated by the input vector is similar to each of the characteristic patterns of the basis vectors is determined. This is given by the modulus squared of the probability amplitudes (P, Q, R, S, T, U, V, W ). The input vector is then classified based on the basis state vector that gives the highest probability value. Figure~\ref{fig:probDistrExamples} shows the probabilities associated with the random input in Figure~\ref{fig:ExamplesInputTransf}. It is obvious that this input is most similar to state basis vector $|100\rangle$ as previously observed. 

Subject to this insight, the proposed IQFT-inspired algorithm for RGB image segmentation  is presented in Algorithm\ref{algorithm1},  where the input $P_m$ is a 3D vector of RGB intensities of the $m$th pixel, $T$ is the total number of pixels, $W$ is a complex 8 by 8 matrix in equation~(\ref{eq:QIFT13}), $\theta_1$, $\theta_2$, and $\theta_3$ are angle parameters for transforming  pixel intensities into phase values, and the output is the required pixel label $l_m {\in} \{0,1,2,..., 7\}$. The segmentation algorithm involves a normalization process in Line 1, linear transformation in Line 2, dimensional transformation from 3D to 8D vector in Line 3, and probability measure in Line 4. A pixel is classified according to the basis vector with the highest probability.

\floatstyle{ruled}
\newfloat{algorithm}{htbp}{loa}
\floatname{algorithm}{Algorithm}
\begin{algorithm}
	\caption{: IQFT-inspired algorithm for RGB image segmentation}
	\label{algorithm1}
	\begin{tabbing}
		\= 	\textbf{Input:}  \\
		\>$I=\mathrm{\{}$ $P_m\in {\mathbb{R}}^3\}$, $m=[1,T]$\\
		\>${\theta }_1,\ {\theta }_2,\ {\theta }_3\ \in\mathbb{R}$ \\
		\>$W\in {\mathbb{C}}^{8\times 8}$\\
		\textbf{Output:} $\mathcal{L}\boldsymbol{=}\left\{l_m\boldsymbol{\in}\mathbb{Z}\right\}$ \\
		\textbf{for }m=1\textbf{to} T\textbf{ do}\\
		\>1. $\left\{P_m\right\}\longleftarrow \left\{P_m/255\right\}, \left\{P_m\right\} =\left\{R_m,G_m,B_m\right\}$\\
		\>2. $\{{\gamma }_m,{\beta }_m,{\alpha }_m\}\longleftarrow\left\{R_m\times{\theta }_1,G_m\times{\theta }_2,B_m\times\ {\theta }_3\right\}$\\
		\>3. $\left\{F_m\right\}=\left\{ \begin{array}{c}
			1 \\
			e^{i\gamma } \\
			e^{i\beta } \\
			e^{i\left(\beta +\gamma \right)} \\
			e^{i\alpha } \\
			e^{i\left(\alpha +\gamma \right)} \\
			e^{i\left(\alpha +\beta \right)} \\
			e^{i\left(\alpha +\beta +\gamma \right)} \end{array}
		\right\}\longleftarrow \left\{ \begin{array}{c}
			{\gamma }_m \\
			{\beta }_m \\
			{\alpha }_m \end{array}\right\}$\\
		\>4. $\{S_m\}\longleftarrow {\left[abs\left(Dot\ Product\left(F_m,W\right)/8\right)\right]}^2\ $\\ 
		\>5. $\{l_m\}\longleftarrow \left\{arg{\mathrm{max} \{S_m\}\ }\right\}$ 
	\end{tabbing}
\end{algorithm}

\subsection{Proposed IQFT-inspired Algorithm for Grayscale Image Segmentation} \label{subsec:GrayScale}
\begin{figure}[h!]
	\centering
	\begin{minipage}{0.23\linewidth}
		\centering
		\includegraphics[width=\linewidth]{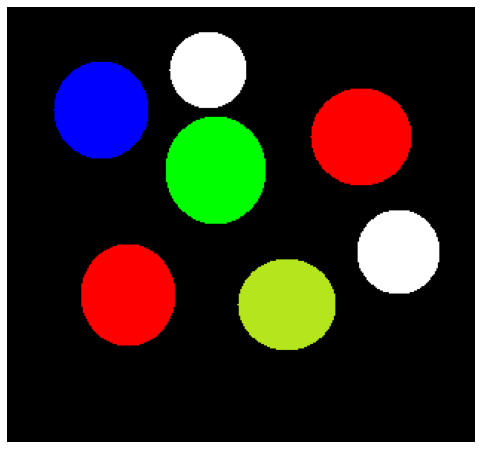} \\
		\caption*{Image}
		\label{ball_original}
	\end{minipage}
	\begin{minipage}{0.23\linewidth}
		\centering
		\includegraphics[width=\linewidth]{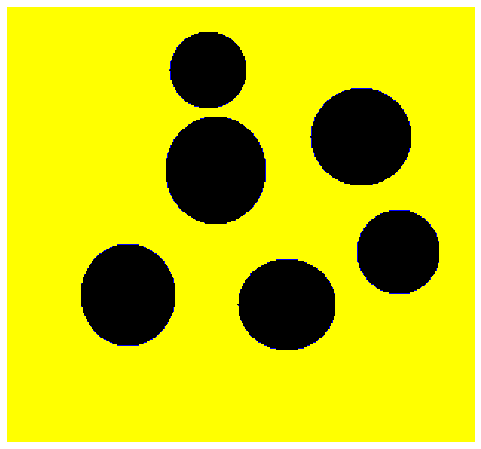} 
		\caption*{K-means}
		\label{ball_kmeans}
	\end{minipage}
	\begin{minipage}{0.23\linewidth}
		\centering
		\includegraphics[width=\linewidth]{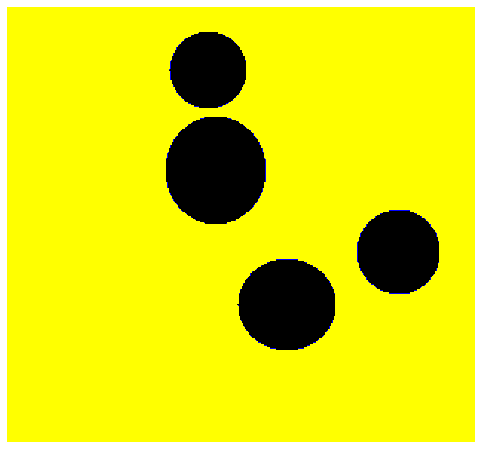} \\
		\caption*{Otsu}
		\label{ball_otsu}
	\end{minipage}
	\begin{minipage}{0.23\linewidth}
		\centering
		\includegraphics[width=\linewidth]{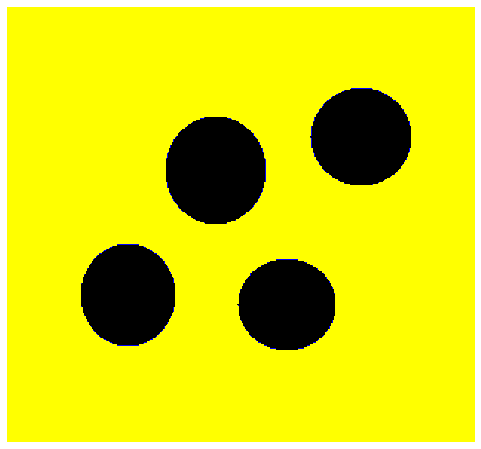} \\
		\caption*{IQFT}
		\label{ball_our_method}
	\end{minipage}
	\caption{Application of multiple thresholding}
	\label{fig:multiple_thresholding}
\end{figure}

Since our approach in Section~\ref{subsec:IQFTInspiredAlgorithm} is not limited by the image color space, it can be adapted for segmentation of grayscale images. In this case, the governing mathematical equation is given by equation~(\ref{eq:IQFTgrayscale1}), where the probability amplitudes, P and Q, are determined from equation~(\ref{eq:IQFTgrayscale2}). For a normalized pixel intensity $I$, and chosen angle parameter $\theta$,  $\gamma=I\theta$. 
\begin{IEEEeqnarray}{l}
	IQF T \left(\frac{1}{\sqrt{2}}\left(|0\rangle+e^{i \gamma}|1\rangle\right)\right) \equiv P|0\rangle+Q|1\rangle
	\label{eq:IQFTgrayscale1}
\end{IEEEeqnarray}
\begin{IEEEeqnarray}{l}
	\left[\begin{array}{c}
		P \\ Q 
	\end{array}\right] 
	\equiv~\frac{1}{2} \times \left[\begin{array}{cc}
		1 & 1 \\
		1 & -1  
	\end{array}\right] \left[\begin{array}{c}
		1 \\
		e^{i \gamma} 
	\end{array}\right]
	\label{eq:IQFTgrayscale2}
\end{IEEEeqnarray}
 In this grayscale implementation, a pixel can be classified into one of two classes based on the probabilities given by equation~(\ref{eq:prob1_prob2}).  For $P(class1)=P(class 2)$, $I=I_{th}$, where $I_{th}$ is a threshold value given by equation~(\ref{eq:threshold}), where $k\in\mathbb{Z}$. 
 \begin{equation}
 	\begin{aligned}
 		\mathrm{p}\left(\mathrm{class1}\right)&=&\frac{\left(\mathrm{1+}\mathrm{cos}{\left(\mathrm{I\theta}\right)}\right)^\mathrm{2}\mathrm{+}\ {\mathrm{(}\mathrm{sin}{\mathrm{(I\theta))}}}^\mathrm{2}}{\mathrm{4}}\\
 		\mathrm{p}\left(\mathrm{class2}\right)&=&\frac{\left(\mathrm{1-}\mathrm{cos}{\left(\mathrm{I\theta}\right)}\right)^\mathrm{2}\mathrm{+}\ {\mathrm{(}\mathrm{sin}{\mathrm{(I\theta))}}}^\mathrm{2}}{\mathrm{4}}
 		\label{eq:prob1_prob2}
 	\end{aligned}
 \end{equation}
 \begin{equation}
 	cos I_{th}\theta=0 \Longrightarrow I_{th}\theta=\left(4k\pm1\right)\frac{\pi}{2\theta}\le1
 	\label{eq:threshold}
 \end{equation}
 Therefore, selecting a particular value of $\theta$ is equivalent to setting a threshold value, and the network behaves like a thresholding technique. Table~\ref{tab:angle_thresholding} shows some $\theta$ values and the corresponding threshold using equation~(\ref{eq:threshold}). Considering equation~(\ref{eq:threshold}), with a single selection of $\theta$, it is possible to establish multiple thresholds. For instance, choosing $\theta=4\pi$ results in four thresholds as shown in equation~(\ref{eq:4thresholds}) for $k=0,1,2$, respectively. The advantage of these multiple thresholds is captured in the  following example. Consider the task of separating red , green, and lemon balls from the others of lower and higher intensities in Figure~\ref{fig:multiple_thresholding}. To achieve the desired goal, $\theta$ is simply set to $4\pi$. However, Otsu-thresholding would require two clearly defined thresholds to achieve the set goal. 
\begin{equation}
	I_{th}=\left(4k\pm1\right)\frac{\pi}{2\theta}=\frac{1}{8},\ \frac{3}{8},\ \frac{5}{8},\ \frac{7}{8}
	\label{eq:4thresholds}
\end{equation}

\begin{table}[htbp]
	\centering
	\caption{Parameter $\theta$ and the corresponding threshold value using equation~\ref{eq:threshold}.}
	\begin{tabular}{|c|c|} \hline
	\textbf{Parameter, $\theta $} &\textbf{Threshold value, $I_{th}$ }\\ \hline
	$3\pi /4$ &0.667 \\ \hline
	$\pi $ & 0.500 \\ \hline
	$5\pi /4$ &0.400 \\ \hline
	$3\pi /2$ &0.333 \\ \hline
	$7\pi /4$ &0.285, 0.857 (multiple) \\ \hline
	$2\pi $ &0.25, 0.75 (multiple)  \\ \hline
	\end{tabular}
	\label{tab:angle_thresholding}
\end{table}

\section{Results and Analysis}\label{sec:results}
\subsection{Dataset}\label{subsec:Dataset}
Considering that most image dataset exist in the classical domain, we have performed simulations of the proposed method in the classical domain.
To demonstrate the effectiveness of our proposed method for Image segmentation, we conduct several experiments on two datasets: (1) the training and validation dataset of PASCAL VOC 2012 segmentation benchmark~\cite{Unsupervised_learning_of_image_segmentation_based_on_differentiable_feature_clustering_kim2020}, focusing only on the segmentation category which contains 2913 labeled set; (2) the xVIEW2 challenge dataset~\cite{xbd_A_dataset_for_assessing_building_damage_from_satellite_imagery_gupta2019} focussing on the 148 RGB satellite pre-disaster images for a ``joplin-tornado" disaster.  

\subsection{Experimental Setup}\label{subsec:ExpSetup}
All images are RGB, and the corresponding grayscale images are prepared by calculating the weighted sum of the corresponding red, green and blue pixels according to Scikit-image~\cite{scikit_image_mage_processing_in_Python_van_2014}, using equation~(\ref{equ: Scikit_image}).
\begin{equation}
	Y = R \times 0.2125 + G \times 0.7154 + B \times 0.0721
	\label{equ: Scikit_image}
\end{equation}
 K-means~\cite{Classification_and_analysis_of_multivariate_observations_macqueen1967}  and Otsu-thresholding~\cite{A_threshold_selection_method_from_graylevel_histograms_otsu1979} are selected as the baseline methods for performance comparison with our method. To implement the baseline methods, we used the scikit-learn library~\cite{Scikit_learn_Machine_Learning_in_Python_Pedregosa_2011} with default settings  for K-means, and scikit-image library~\cite{scikit_image_mage_processing_in_Python_van_2014} for  Otsu-thresholding.  All the algorithms used in this study are coded using Python on MacBook Pro with 8-Core Intel Core i9 running at 2.3 GHz.

\subsection{Evaluation Metric}\label{subsec:EvalMetric}
The segmentation accuracies of the methods used in this study are assessed by the mean intersection over union (mIOU) score defined in equation~(\ref{eq:mIOU}) and equation~(\ref{eq:IOU}), where TP, FP, FN, T, and P are true positive, false positive, false negative, ground truth and prediction, respectively. TensorFlow function~\cite{LargeScale_Machine_Learning_on_Heterogeneous_Systems_tensorflow2015}, utilizing equation~(\ref{eq:mIOU}), is used for all mIOU calculations. Pixels  around the border of an object that are marked ‘void’ in the ground truth are not used in our calculations \cite{The_pascal_visual_object_classes_challenge_A_retrospective_everingham2015}. 
\begin{equation}
	mIOU=\frac{IOU(foreground)+IOU(background)}{2}
	\label{eq:mIOU}
\end{equation}
\begin{equation}
	IOU=\frac{TP}{TP+FP+FN}=\frac{T \cup P}{T \cap P}
	\label{eq:IOU}
\end{equation}

\subsection{Experimental Results}\label{subsec:ExperimentalResult}
For the purpose of evaluating the effects of design choices on performance, different experiments are conducted on standard datasets and the results obtained are presented in this section.

\subsubsection{Effect of the normalization process}\label{subsec:NormProcess}
\begin{figure}[h!]
	\centering
	\begin{minipage}{0.45\linewidth}
		\centering
		\includegraphics[height=0.1\textheight]{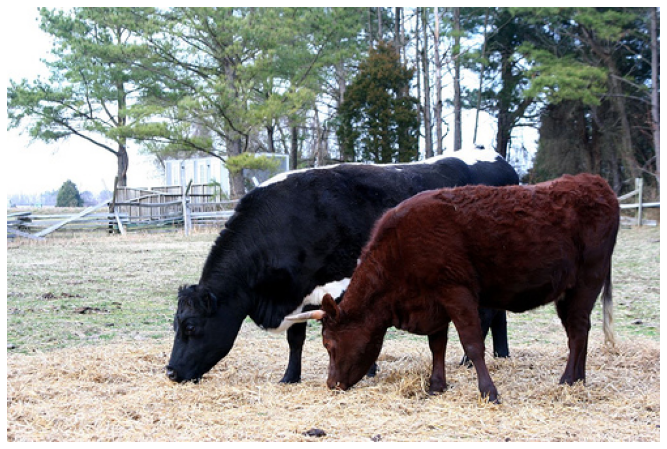} \\
		\includegraphics[height=0.1\textheight]{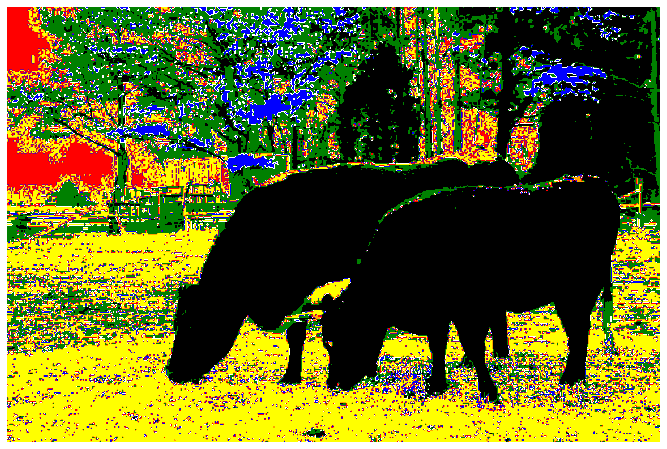} \\
		\includegraphics[height=0.1\textheight]{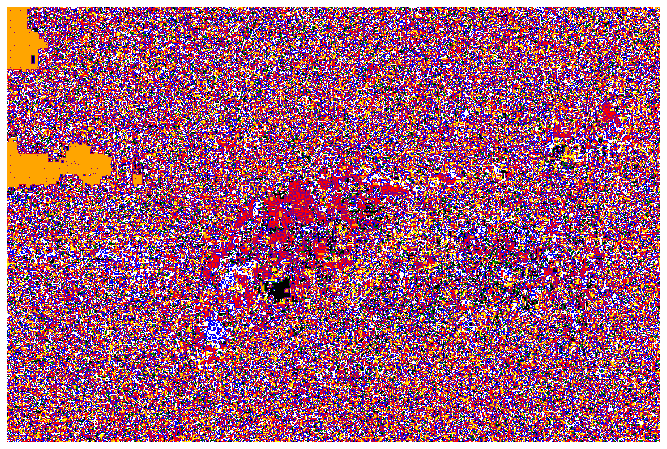} \\
	\end{minipage}
	\begin{minipage}{0.45\linewidth}
		\centering
		\includegraphics[height=0.1\textheight]{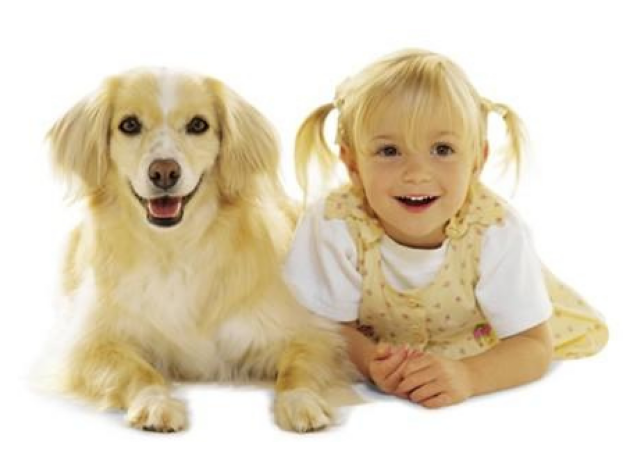} \\
		\includegraphics[height=0.1\textheight]{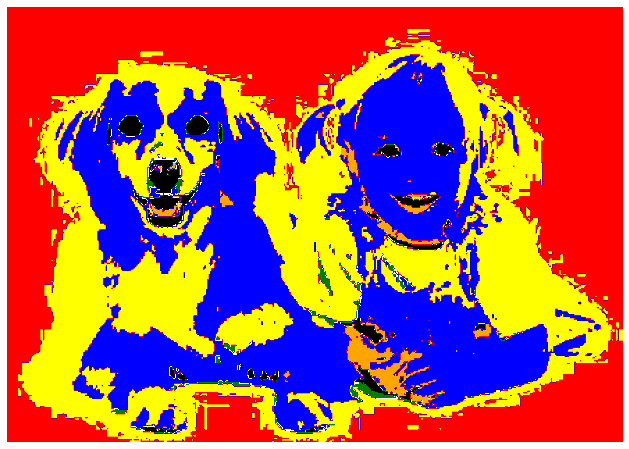} \\
		\includegraphics[height=0.1\textheight]{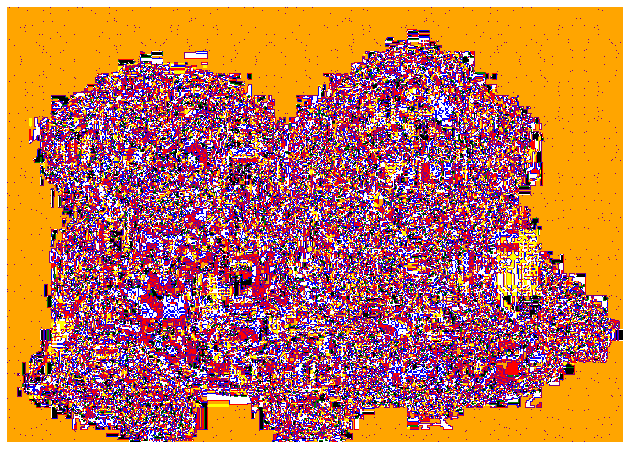} \\
	\end{minipage}
	\caption{Effect of the normalization process. First row contains two different images, second rows contains  segmentation patterns when the normalization process is included, and the last row shows segmentation patterns when the normalization process is not included}
	\label{fig:NormProcess}
\end{figure}

Here, the effect of the normalization process of Section~\ref{subsec:IQFTInspiredAlgorithm} on the quality of the segmentation pattern is investigated. The results obtained in Figure~\ref{fig:NormProcess} show that, for a smooth segmentation pattern, normalized image intensities are required to avoid ``noisy'' segments. \vspace{1ex}

\subsubsection{Effects of  angle parameter ($\theta_1, \theta_2, \theta_3$) on the number of segments}
\label{subsec:NumSegRGB}
\begin{figure}[h!]
	\centering
	\begin{minipage}{0.32\linewidth}
		\centering
		\includegraphics[height=0.06\textheight]{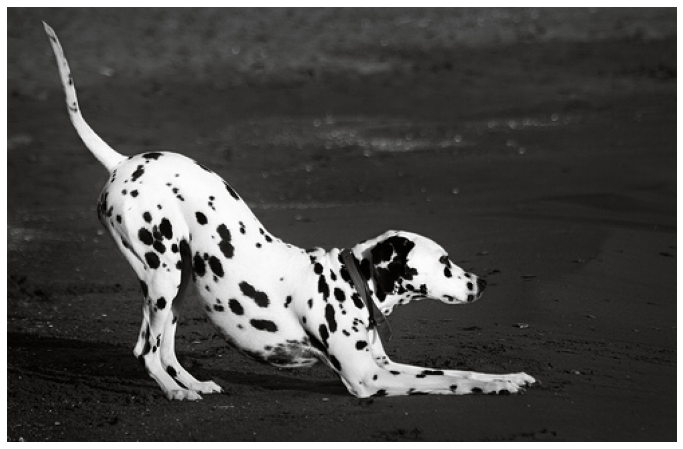} \\
		\caption*{Image}
	\end{minipage}
	\begin{minipage}{0.32\linewidth}
		\centering
		\includegraphics[height=0.06\textheight]{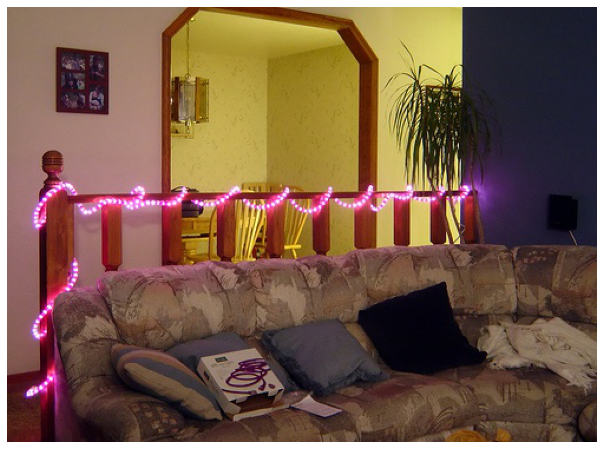} \\
		\caption*{Image}
	\end{minipage}
	\begin{minipage}{0.32\linewidth}
		\centering
		\includegraphics[height=0.06\textheight]{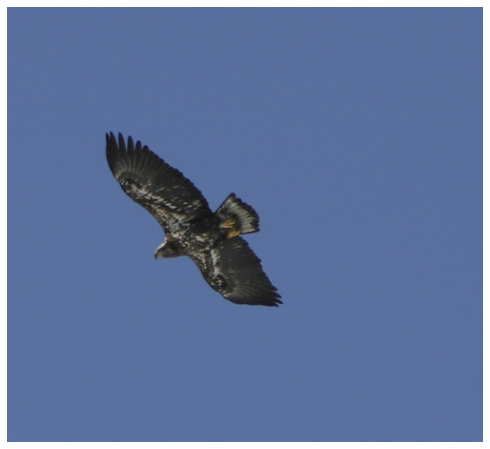} \\
		\caption*{Image}
	\end{minipage}
	\begin{minipage}{0.32\linewidth}
		\centering
		\includegraphics[height=0.06\textheight]{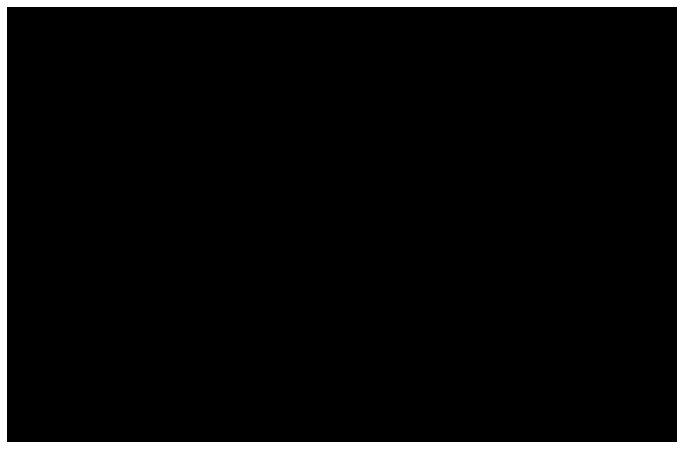} \\
		\caption*{$\theta=\pi/4$, 1-seg}
	\end{minipage}
	\begin{minipage}{0.32\linewidth}
		\centering
		\includegraphics[height=0.06\textheight]{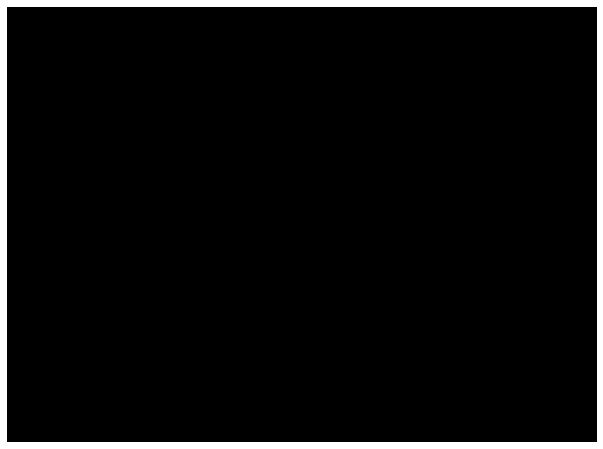} \\
		\caption*{$\theta=\pi/4$, 1-seg.}
	\end{minipage}
	\begin{minipage}{0.32\linewidth}
		\centering
		\includegraphics[height=0.06\textheight]{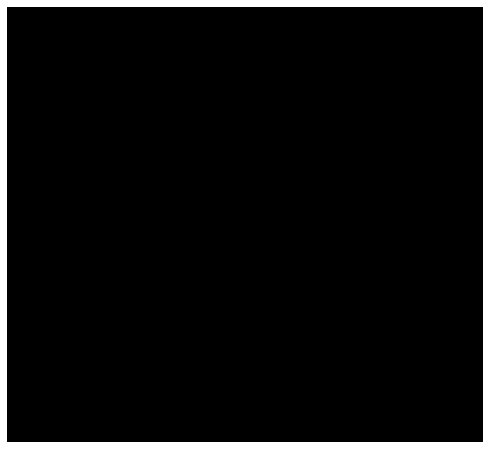} \\
		\caption*{$\theta=\pi/4$, 1-seg.}
	\end{minipage}
	\begin{minipage}{0.32\linewidth}
		\centering
		\includegraphics[height=0.06\textheight]{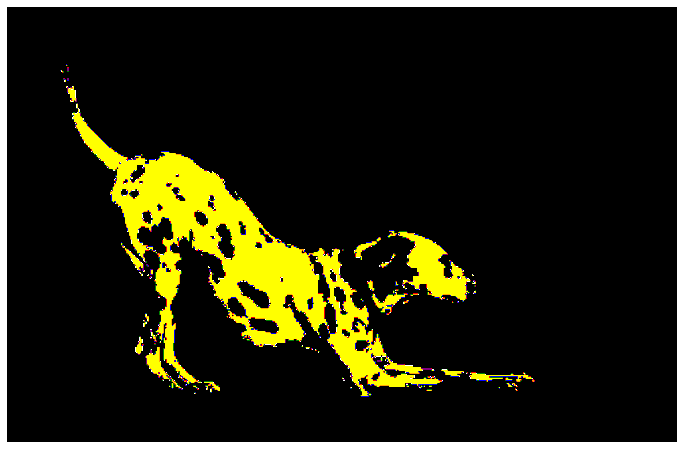} \\
		\caption*{$\theta=\pi/2$, 2-seg.}
	\end{minipage}
	\begin{minipage}{0.32\linewidth}
		\centering
		\includegraphics[height=0.06\textheight]{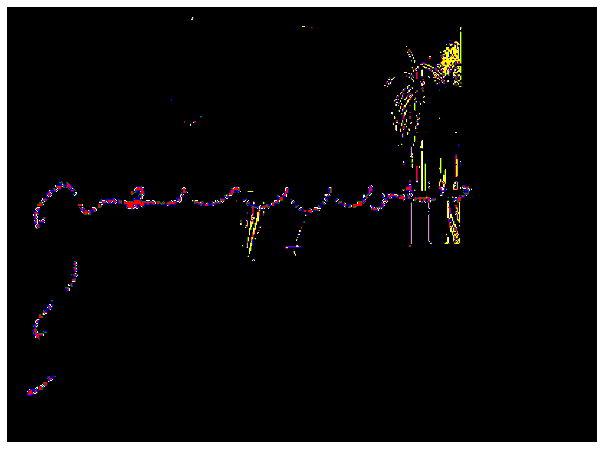} \\
		\caption*{$\theta=\pi/2$, 3-seg.}
	\end{minipage}
	\begin{minipage}{0.32\linewidth}
		\centering
		\includegraphics[height=0.06\textheight]{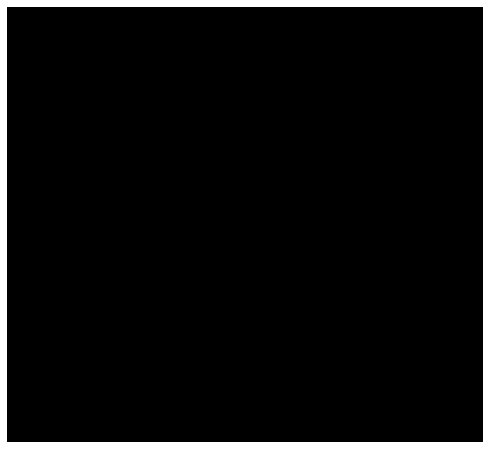} \\
		\caption*{$\theta=\pi/2$, 1-seg.}
	\end{minipage}
	\begin{minipage}{0.32\linewidth}
		\centering
		\includegraphics[height=0.06\textheight]{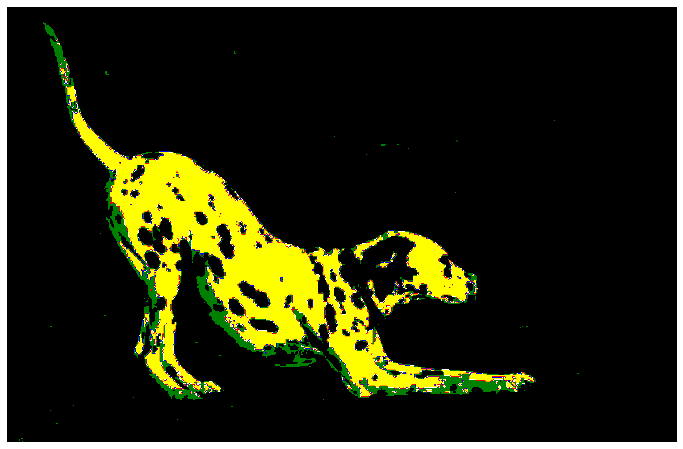} \\
		\caption*{$\theta=\pi$, 4-seg.}
	\end{minipage}
	\begin{minipage}{0.32\linewidth}
		\centering
		\includegraphics[height=0.06\textheight]{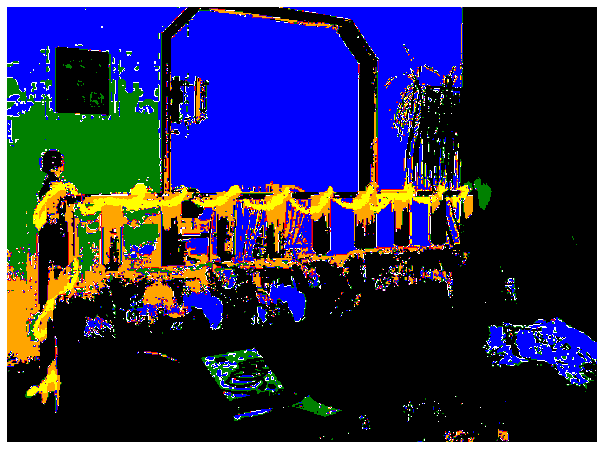} \\
		\caption*{$\theta=\pi$, 6-seg.}
	\end{minipage}
	\begin{minipage}{0.32\linewidth}
		\centering
		\includegraphics[height=0.06\textheight]{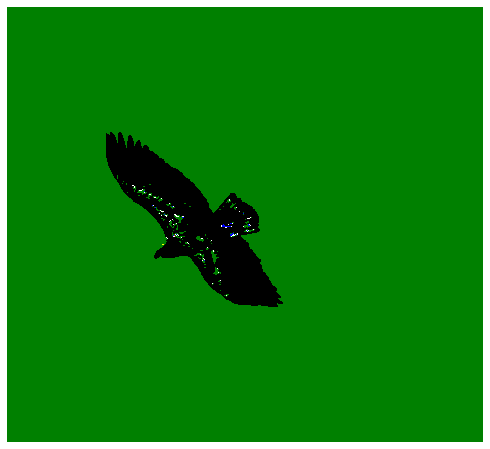} \\
		\caption*{$\theta=\pi$, 5-seg.}
	\end{minipage}
	\begin{minipage}{0.32\linewidth}
		\centering
		\includegraphics[height=0.06\textheight]{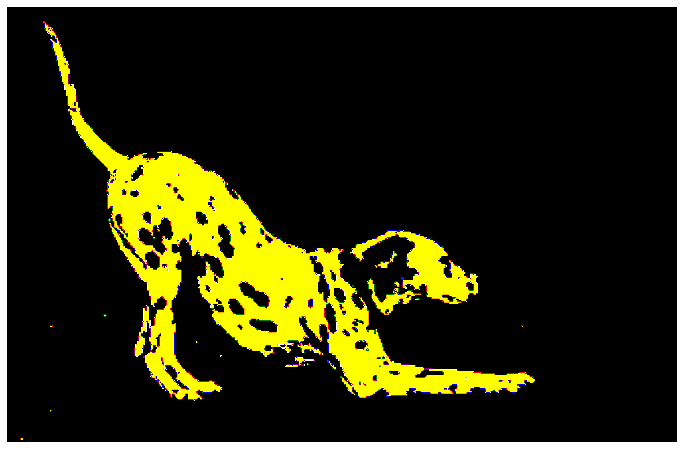} \\
		\caption*{mixed, 2-seg.}
	\end{minipage}
	\begin{minipage}{0.32\linewidth}
		\centering
		\includegraphics[height=0.06\textheight]{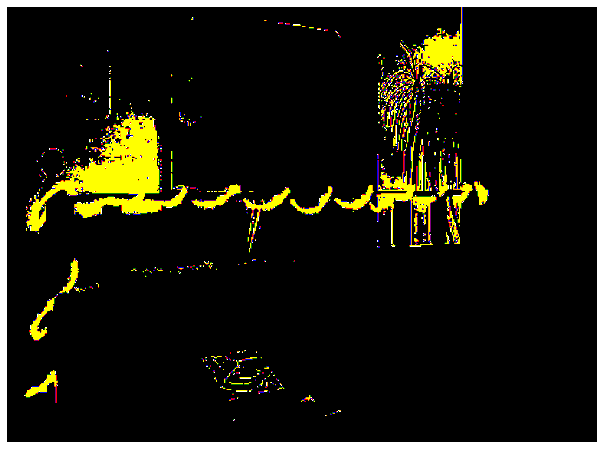} \\
		\caption*{mixed, 2-seg.}
	\end{minipage}
	\begin{minipage}{0.32\linewidth}
		\centering
		\includegraphics[height=0.06\textheight]{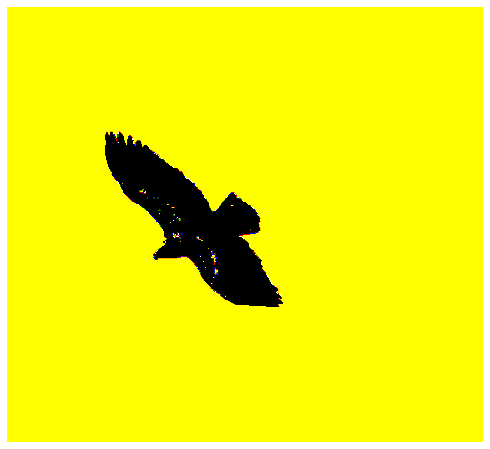} \\
		\caption*{mixed, 2-seg.}
	\end{minipage}
	\caption{Effects of $\theta$ on the number of segments and the segmentation quality--where 'mixed'  in the last row represents $\theta_1=\pi/4, \theta_2=\pi/2, \theta_3= \pi$}
	\label{fig:NumberSeg}
\end{figure}

To study the effects of the angular parameter, $\theta$, on the segmentation pattern, we generated $100,000\times3$ random numbers between $0$ and $1$ as normalized RGB values and determined the maximum number of segments using all the combinations for different values of $\theta$. The results are shown in Table~\ref{tab:NumSegments}. It is evident that the number of segments varies with the angular parameter. This is due to the modification of the transformed input pattern as the angular parameters are varied. This effect is observed on real images shown in Figure~\ref{fig:NumberSeg}.
For $\theta_1=\theta_2=\theta_3=\pi/4$, the six arguments from the complex representation of each pixel are located between 0 and $3\pi/4$ radian. This tends to produce a pattern most similar to the $|000\rangle$ state. Therefore, all pixels are classified into one segment. For $\theta_1=\theta_2=\theta_3=\pi/2$, the segmentation process is biased towards the low-luminous pixels, as pixels (objects) with strong brightness are visible in Figure~\ref{fig:NumberSeg}. Using different angles can result in some peculiar segmentation effects. This is shown by $\theta_1=\pi/4, \theta_2=\pi/2, \theta_3= \pi$  which usually outputs two segments.  \vspace{3ex}

\begin{table}[htbp]
	\centering
	\caption{Parameter $\theta$ and the possible number of segments}
	\begin{tabular}{|l|c|} \hline
		\textbf{Parameter, $\theta $} &\textbf{max. number of segments}\\ \hline
		$\theta_1=\theta_2=\theta_3= \pi /4$ &1 \\ \hline
		$\theta_1=\theta_2=\theta_3= \pi /2$ & 3 \\ \hline
		$\theta_1=\theta_2=\theta_3= 3\pi /4$ &5 \\ \hline
		$\theta_1=\theta_2=\theta_3= \pi $ &6 \\ \hline
		$\theta_1=\theta_2=\theta_3= 5\pi /4$&8 \\ \hline
		$\theta_1=\theta_2=\theta_3= 3\pi /2$ &8  \\ \hline
		$\theta_1=\theta_2=\theta_3= 7\pi /4$ &8 \\ \hline
		$\theta_1=\theta_2=\theta_3= 2\pi $ &8 \\ \hline
		$\theta_1=\pi/4, \theta_2=\pi/2, \theta_3= \pi $ &2 (constant)  \\ \hline
	\end{tabular}
	\label{tab:NumSegments}
\end{table}

\subsection{Performance comparison}\label{subsec:PerfComp}
\subsubsection{A performance comparison of the IQFT-inspired algorithm for grayscale Images and Otsu-threshold method}
\label{subsec:PerfCompGray}
\begin{figure}[h!]
	\centering
	\begin{minipage}{0.32\linewidth}
		\centering
		\includegraphics[height=0.08\textheight]{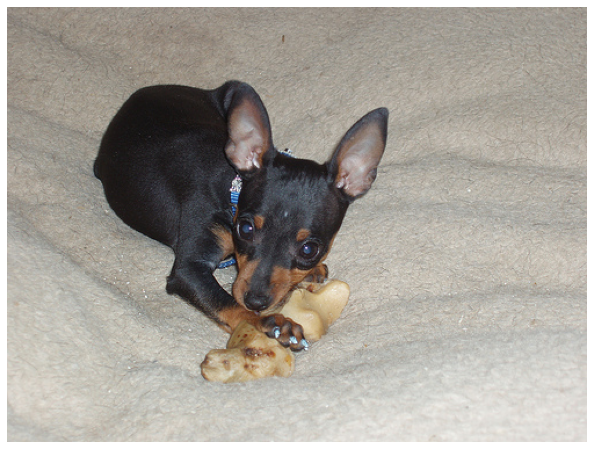} \\
		\caption*{}
	\end{minipage}
	\begin{minipage}{0.32\linewidth}
		\centering
		\includegraphics[height=0.08\textheight]{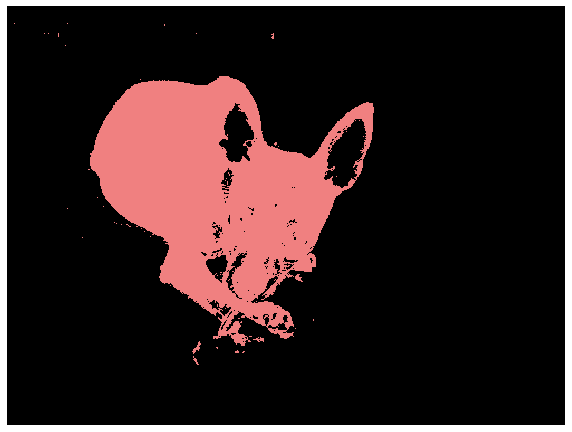} \\
		\caption*{$I_{th}=0.4465$}
	\end{minipage}
	\begin{minipage}{0.32\linewidth}
		\centering
		\includegraphics[height=0.08\textheight]{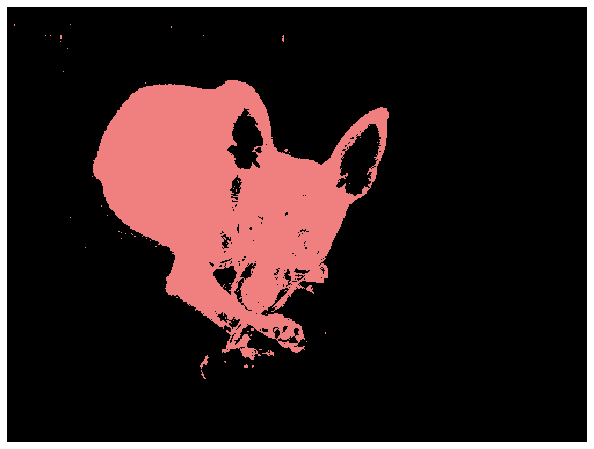} \\
		\caption*{$\theta=1.1197\pi$}
	\end{minipage}
	\begin{minipage}{0.32\linewidth}
	\centering
	\includegraphics[height=0.08\textheight]{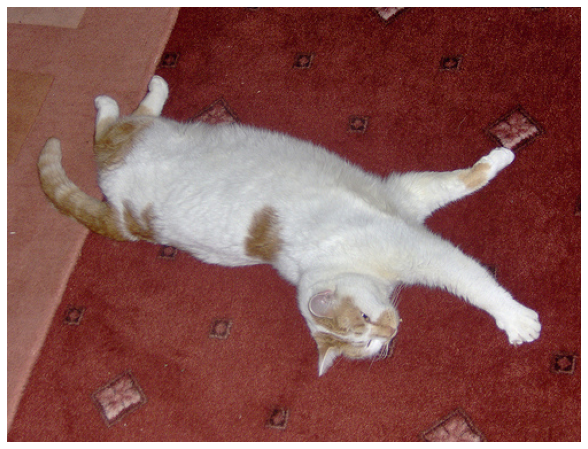} \\
	\caption*{}
	\end{minipage}
	\begin{minipage}{0.32\linewidth}
	\centering
	\includegraphics[height=0.08\textheight]{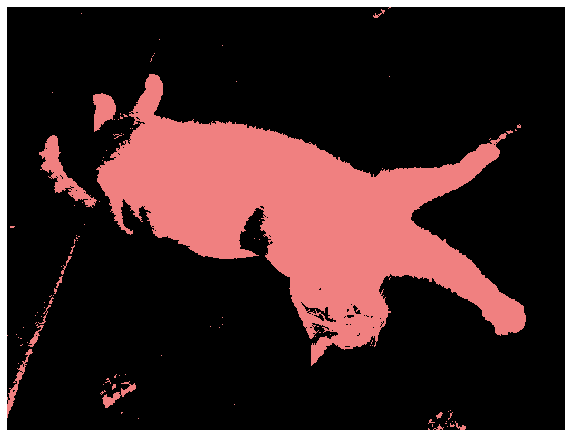} \\
	\caption*{$I_{th}=0.4911$}
	\end{minipage}
	\begin{minipage}{0.32\linewidth}
	\centering
	\includegraphics[height=0.08\textheight]{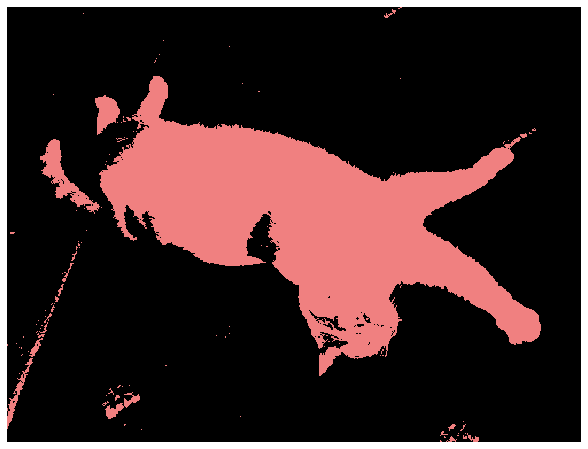} \\
	\caption*{$\theta=1.0180\pi$}
	\end{minipage}
	\caption{Performance comparison of the IQFT-inspired algorithm for grayscale images and Otsu-thresholding. If the threshold from Otsu method is converted to parameter $\theta$ according to equation~(\ref{eq:threshold}), the outputs of the two methods are identical}
	\label{fig:PerfComGray_Otsu}
\end{figure}

As pointed out in Section~\ref{subsec:GrayScale}, setting a value of $\theta$ is equivalent to a threshold value in equation~(\ref{eq:threshold}). To support this conclusion, Figure~\ref{fig:PerfComGray_Otsu} shows two examples of the segmentation results observed from using Otsu-thresholding and  IQFT-inspired algorithm for grayscale images with equivalent angle parameters using equation~(\ref{eq:threshold}).   Therefore, setting $\theta$ according to equation~(\ref{eq:threshold}) results in identical segmentation pattern and equal mIOU values (not shown) for both methods. \vspace{1ex}

\subsubsection{A performance comparison of the proposed method and the baseline methods} \label{subsec:PerfComp}
\begin{figure}[h!]
	\centering
	\begin{minipage}{0.1\linewidth}
		\centering
		\includegraphics[height=0.06\textheight]{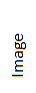} \\
	\end{minipage}
	\begin{minipage}{0.3\linewidth}
		\centering
		\includegraphics[height=0.06\textheight]{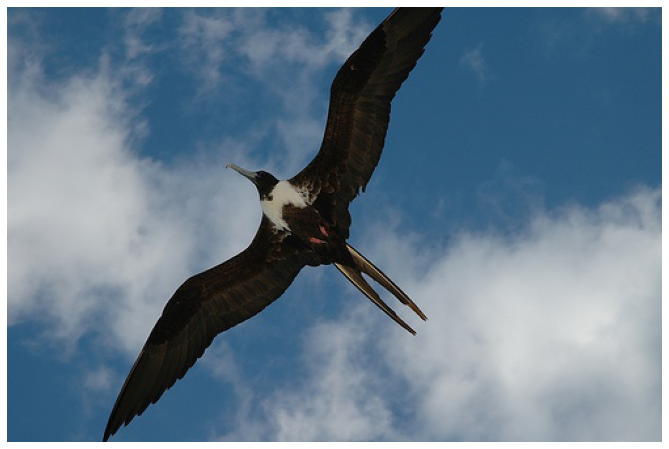} \\
	\end{minipage}
	\begin{minipage}{0.24\linewidth}
		\centering
		\includegraphics[height=0.06\textheight]{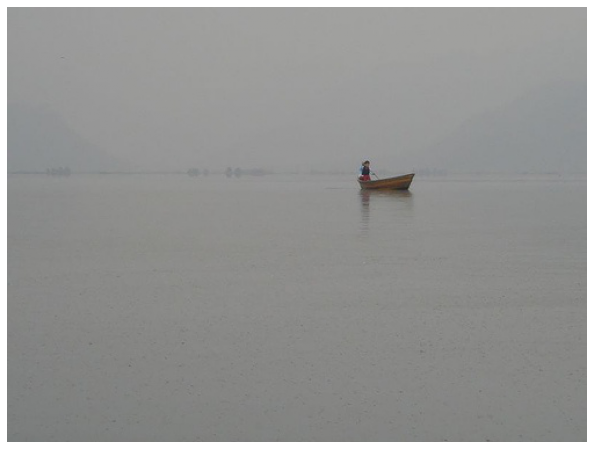} \\
	\end{minipage}
	\begin{minipage}{0.22\linewidth}
		\centering
		\includegraphics[height=0.06\textheight]{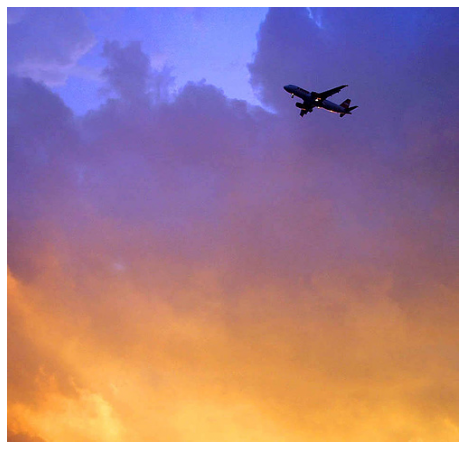} \\
	\end{minipage}
	\begin{minipage}{0.1\linewidth}
		\centering
		\includegraphics[height=0.06\textheight]{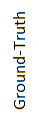} \\
	\end{minipage}
	\begin{minipage}{0.3\linewidth}
		\centering
		\includegraphics[height=0.06\textheight]{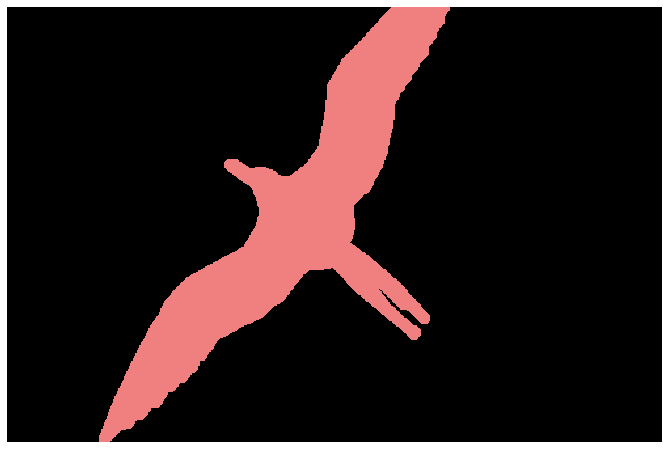} \\
	\end{minipage}
	\begin{minipage}{0.24\linewidth}
		\centering
		\includegraphics[height=0.06\textheight]{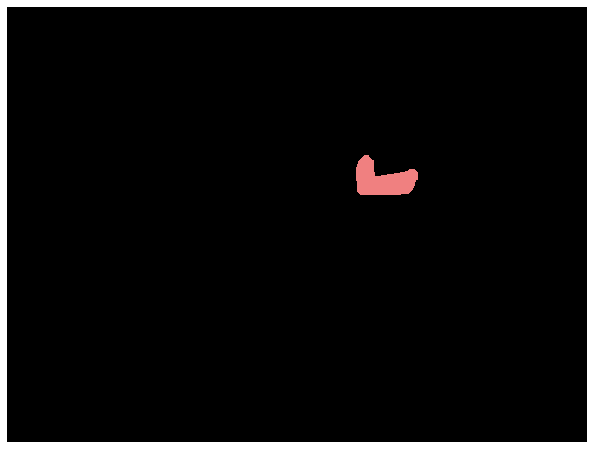} \\
	\end{minipage}
	\begin{minipage}{0.22\linewidth}
	\centering
	\includegraphics[height=0.06\textheight]{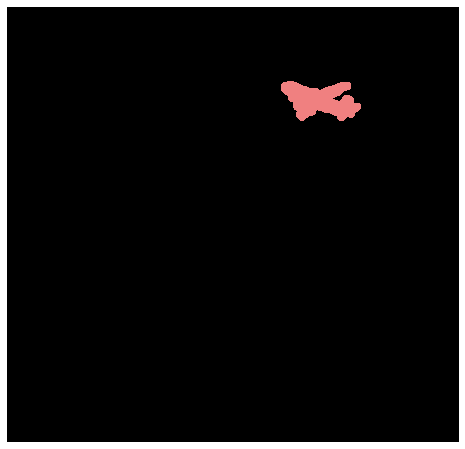} \\
	\end{minipage}
	\begin{minipage}{0.1\linewidth}
		\centering
		\includegraphics[height=0.06\textheight]{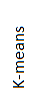} \\
		\caption*{}
	\end{minipage}
	\begin{minipage}{0.3\linewidth}
		\centering
		\includegraphics[height=0.06\textheight]{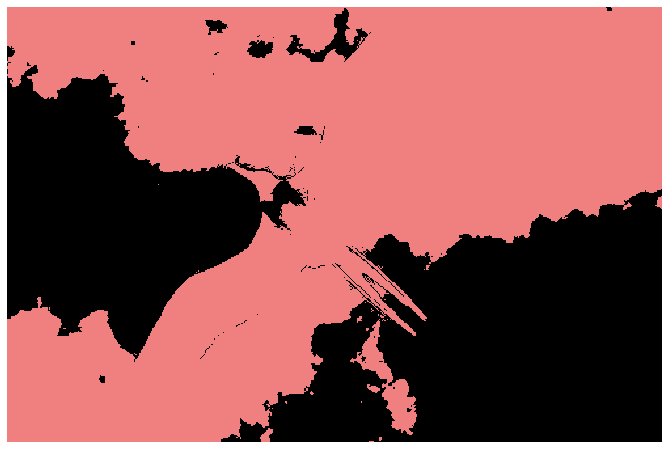} \\
		\caption*{mIOU=0.3197}
	\end{minipage}
	\begin{minipage}{0.24\linewidth}
	\centering
	\includegraphics[height=0.06\textheight]{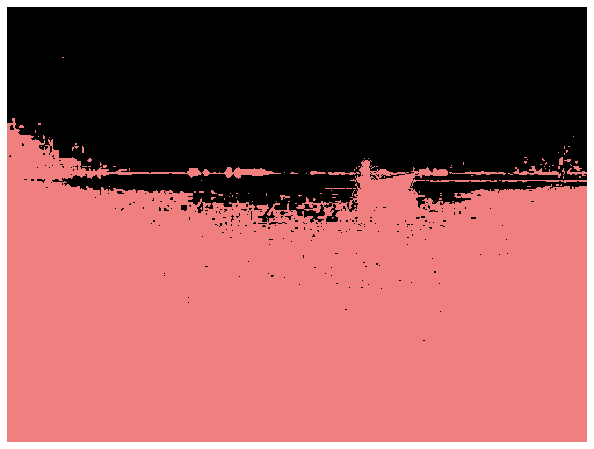} \\
	\caption*{mIOU=0.2080}
	\end{minipage}
	\begin{minipage}{0.22\linewidth}
	\centering
	\includegraphics[height=0.06\textheight]{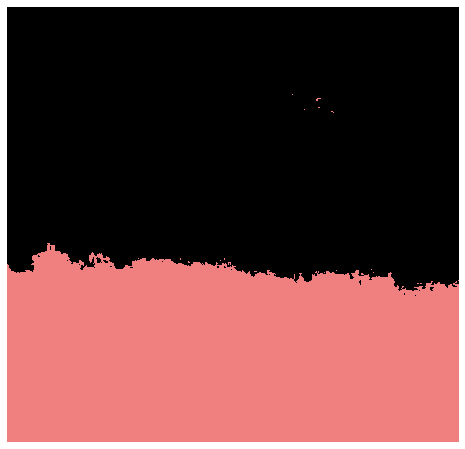} \\
	\caption*{mIOU=0.2036}
	\end{minipage}
	\begin{minipage}{0.1\linewidth}
		\centering
		\includegraphics[height=0.06\textheight]{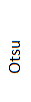} \\
		\caption*{}
	\end{minipage}
	\begin{minipage}{0.3\linewidth}
		\centering
		\includegraphics[height=0.06\textheight]{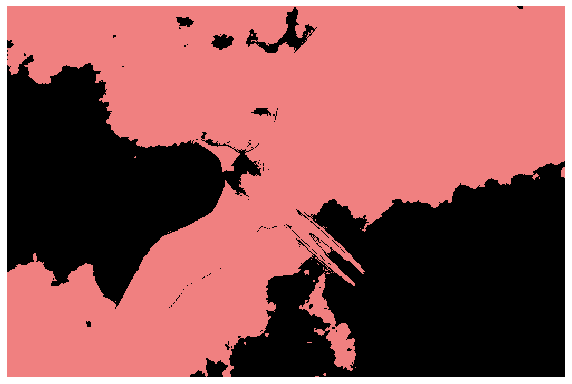} \\
		\caption*{mIOU=0.3179}
	\end{minipage}
	\begin{minipage}{0.24\linewidth}
	\centering
	\includegraphics[height=0.06\textheight]{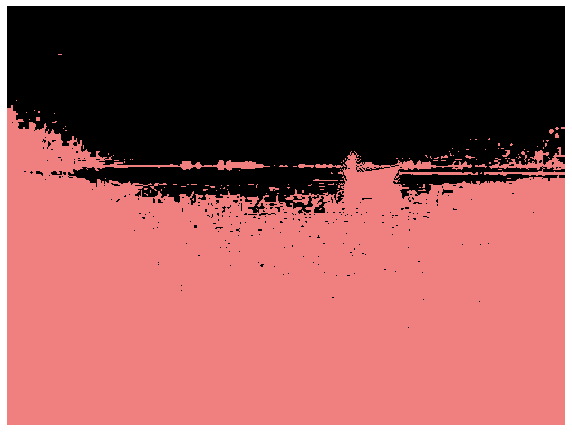} \\
	\caption*{mIOU=0.2052}
	\end{minipage}
	\begin{minipage}{0.22\linewidth}
	\centering
	\includegraphics[height=0.06\textheight]{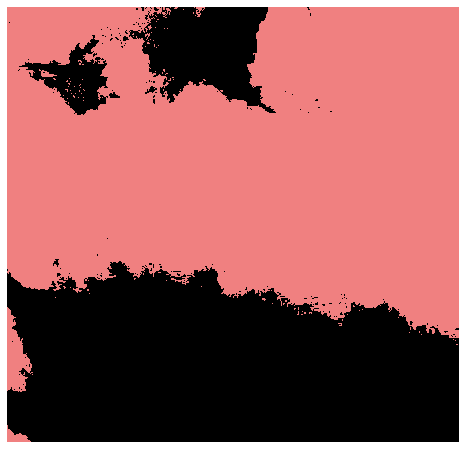} \\
	\caption*{mIOU=0.2036}
	\end{minipage}
	\begin{minipage}{0.1\linewidth}
		\centering
		\includegraphics[height=0.05\textheight]{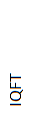} \\
		\caption*{}
	\end{minipage}
	\begin{minipage}{0.3\linewidth}
	\centering
	\includegraphics[height=0.06\textheight]{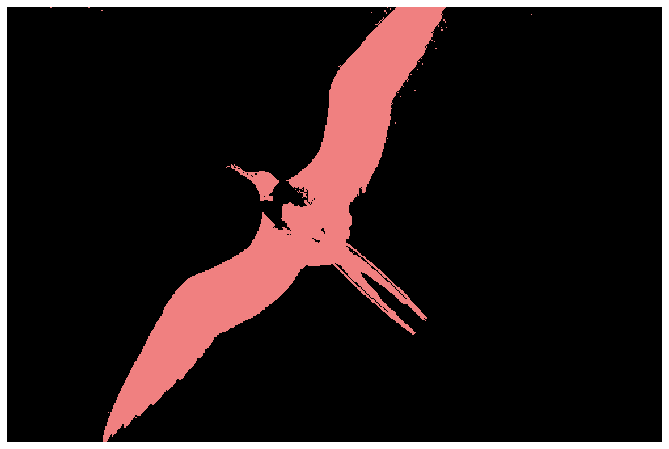} \\
	\caption*{mIOU=0.9762}
	\end{minipage}
	\begin{minipage}{0.24\linewidth}
		\centering
		\includegraphics[height=0.06\textheight]{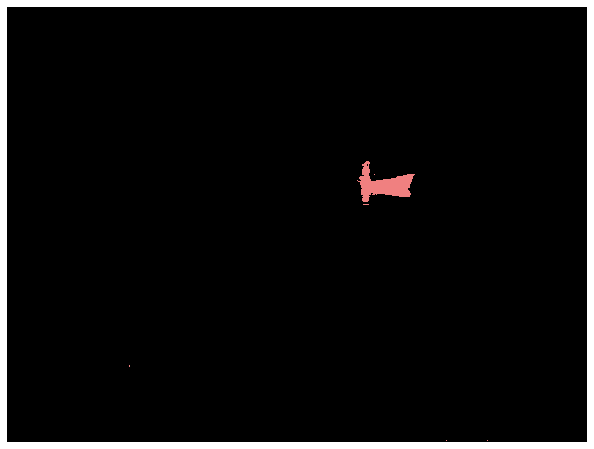} \\
		\caption*{mIOU=0.9187}
	\end{minipage}
	\begin{minipage}{0.22\linewidth}
		\centering
		\includegraphics[height=0.06\textheight]{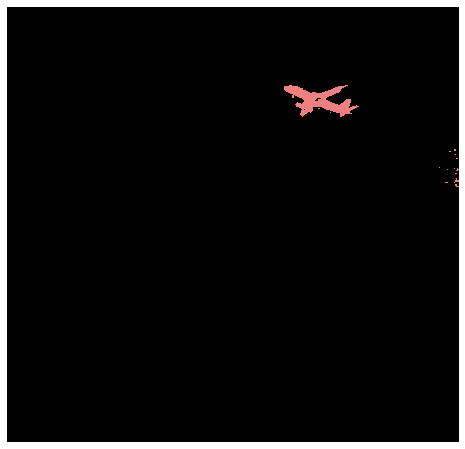} \\
		\caption*{mIOU=0.9697}
	\end{minipage}
	\caption{Results of segmentation with reference images on PASCAL VOC 2012. The IQFT-inspired algorithm for RGB images shows better foreground-background segmentation results as shown by the mIOU scores }
	\label{fig:perform_improve1}
\end{figure}

\begin{figure}[h!]
	\centering
	\begin{minipage}{0.07\linewidth}
		\centering
		\includegraphics[width=\linewidth]{figures/image_image.png} \\
	\end{minipage}
	\begin{minipage}{0.28\linewidth}
		\centering
		\includegraphics[width=\linewidth]{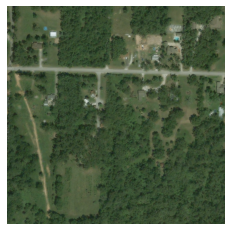} \\
	\end{minipage}
	\begin{minipage}{0.28\linewidth}
		\centering
		\includegraphics[width=\linewidth]{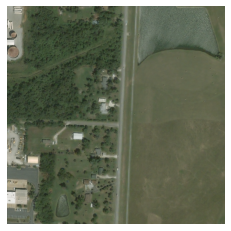} \\
	\end{minipage}
	\begin{minipage}{0.28\linewidth}
		\centering
		\includegraphics[width=\linewidth]{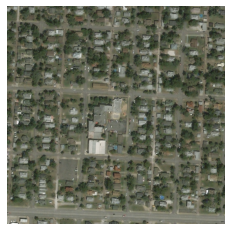} \\
	\end{minipage}
	\begin{minipage}{0.07\linewidth}
		\centering
		\includegraphics[width=\linewidth]{figures/image_GT.png} \\
	\end{minipage}
	\begin{minipage}{0.28\linewidth}
		\centering
		\includegraphics[width=\linewidth]{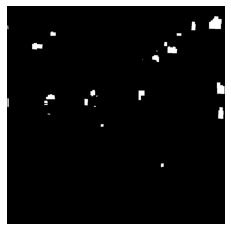} \\
	\end{minipage}
	\begin{minipage}{0.28\linewidth}
		\centering
		\includegraphics[width=\linewidth]{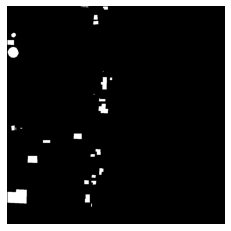} \\
	\end{minipage}
	\begin{minipage}{0.28\linewidth}
		\centering
		\includegraphics[width=\linewidth]{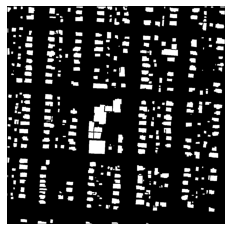} \\
	\end{minipage}
	\begin{minipage}{0.07\linewidth}
		\centering
		\includegraphics[width=\linewidth]{figures/image_kmeans.png} \\
		\caption*{}
	\end{minipage}
	\begin{minipage}{0.28\linewidth}
		\centering
		\includegraphics[width=\linewidth]{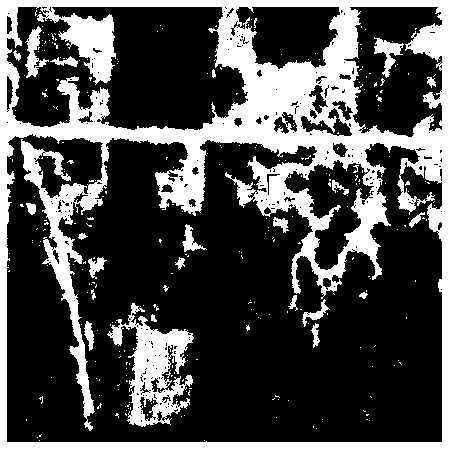} \\
		\caption*{mIOU=0.1175}
	\end{minipage}
	\begin{minipage}{0.28\linewidth}
		\centering
		\includegraphics[width=\linewidth]{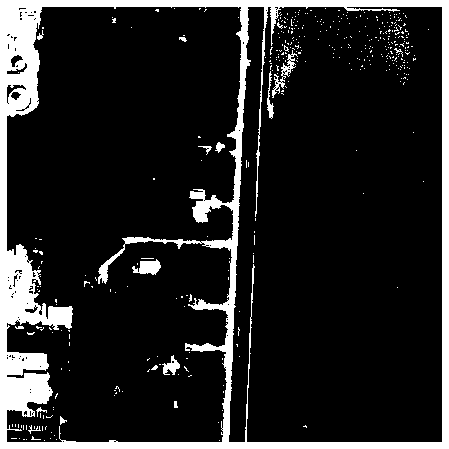} \\
		\caption*{mIOU=0.5243}
	\end{minipage}
	\begin{minipage}{0.28\linewidth}
		\centering
		\includegraphics[width=\linewidth]{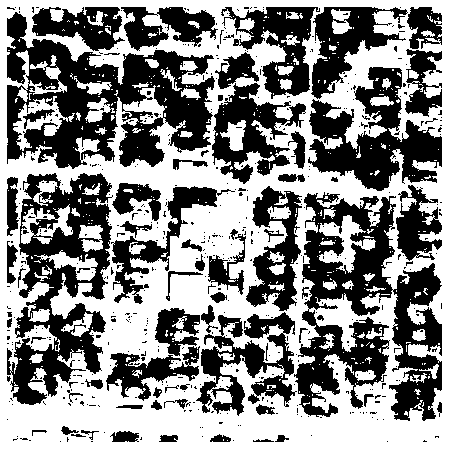} \\
		\caption*{mIOU=0.2394}
	\end{minipage}
	\begin{minipage}{0.07\linewidth}
		\centering
		\includegraphics[width=\linewidth]{figures/image_otsu.png} \\
		\caption*{}
	\end{minipage}
	\begin{minipage}{0.28\linewidth}
		\centering
		\includegraphics[width=\linewidth]{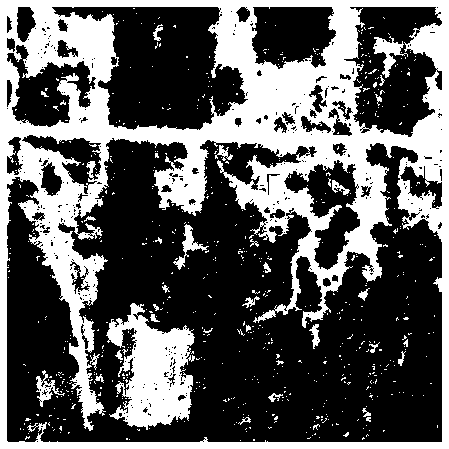} \\
		\caption*{mIOU=0.3597}
	\end{minipage}
	\begin{minipage}{0.28\linewidth}
		\centering
		\includegraphics[width=\linewidth]{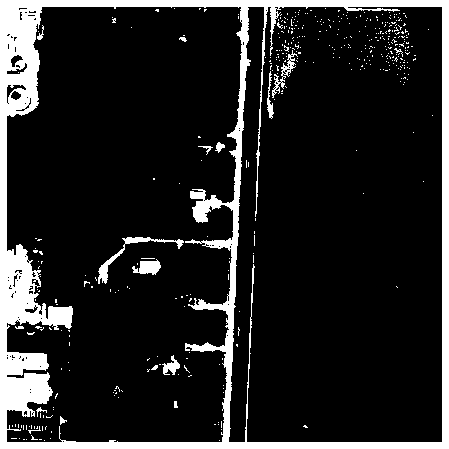} \\
		\caption*{mIOU=0.5226}
	\end{minipage}
	\begin{minipage}{0.28\linewidth}
		\centering
		\includegraphics[width=\linewidth]{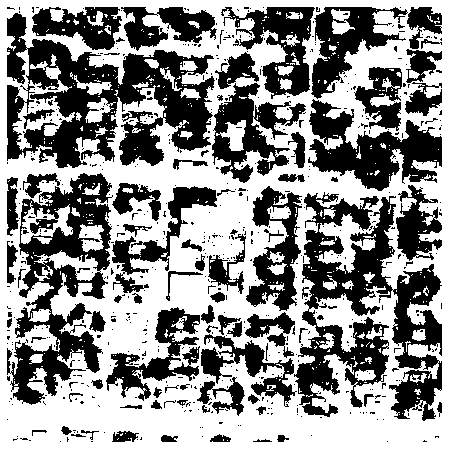} \\
		\caption*{mIOU=0.3390}
	\end{minipage}
	\begin{minipage}{0.07\linewidth}
		\centering
		\includegraphics[width=\linewidth]{figures/image_iqft.png} \\
		\caption*{}
	\end{minipage}
	\begin{minipage}{0.28\linewidth}
		\centering
		\includegraphics[width=\linewidth]{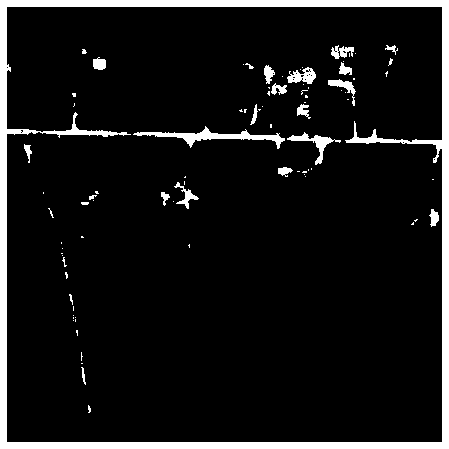} \\
		\caption*{mIOU=0.51295}
	\end{minipage}
	\begin{minipage}{0.28\linewidth}
		\centering
		\includegraphics[width=\linewidth]{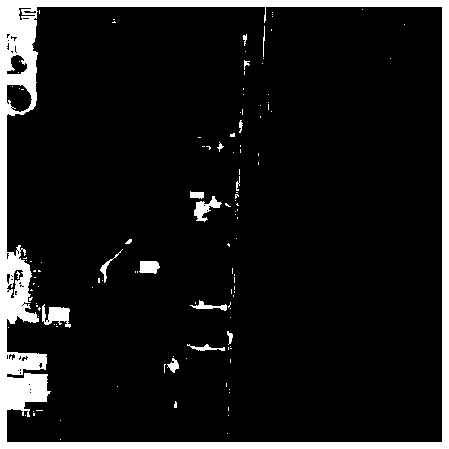} \\
		\caption*{mIOU=0.5683}
	\end{minipage}
	\begin{minipage}{0.28\linewidth}
		\centering
		\includegraphics[width=\linewidth]{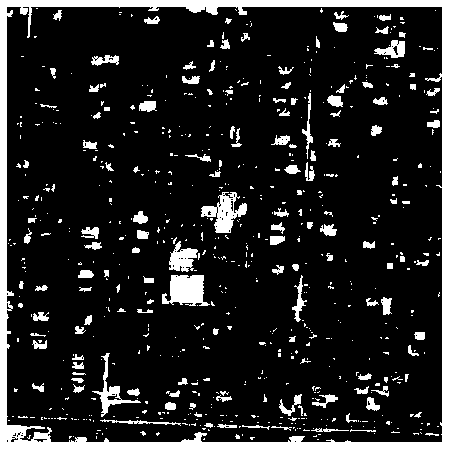} \\
		\caption*{mIOU=0.5453}
	\end{minipage}
	\caption{Segmentation result using xVIEW2 challenge dataset. Measured by the mIOU scores, the IQFT-inspired algorithm for RGB images shows better foreground-background segmentation results.}
	\label{fig:perform_xview}
\end{figure}

\begin{figure}[h!]
	\centering
	\begin{minipage}{0.32\linewidth}
		\centering
		\caption*{Image}
		\includegraphics[height=0.08\textheight]{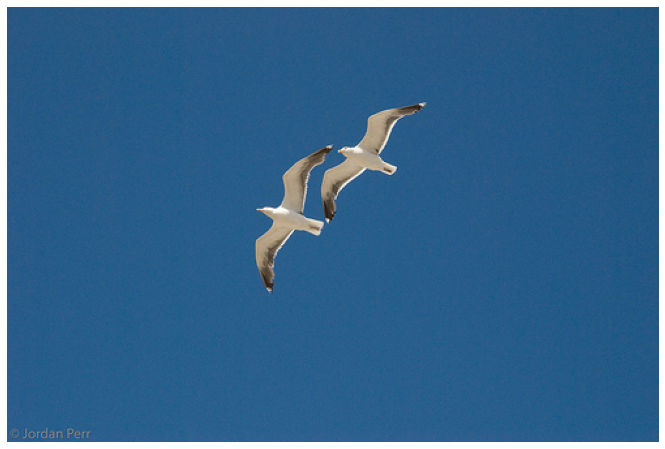} \\
		\caption*{}
	\end{minipage}
	\begin{minipage}{0.32\linewidth}
		\centering
		\caption*{Ground truth}
		\includegraphics[height=0.08\textheight]{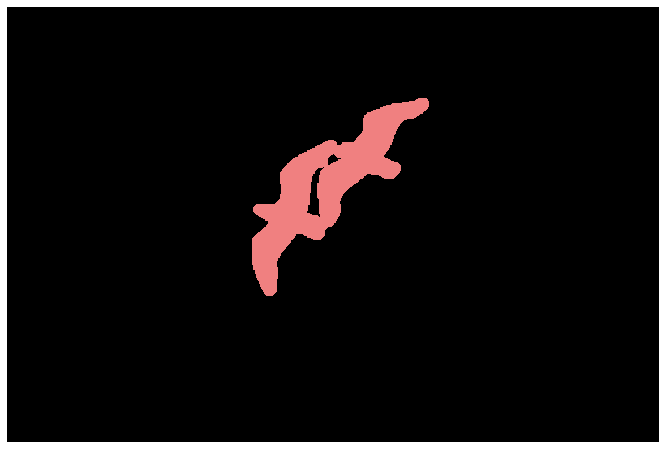} \\
		\caption*{}
	\end{minipage}
	\begin{minipage}{0.32\linewidth}
		\centering
		\caption*{K-means}
		\includegraphics[height=0.08\textheight]{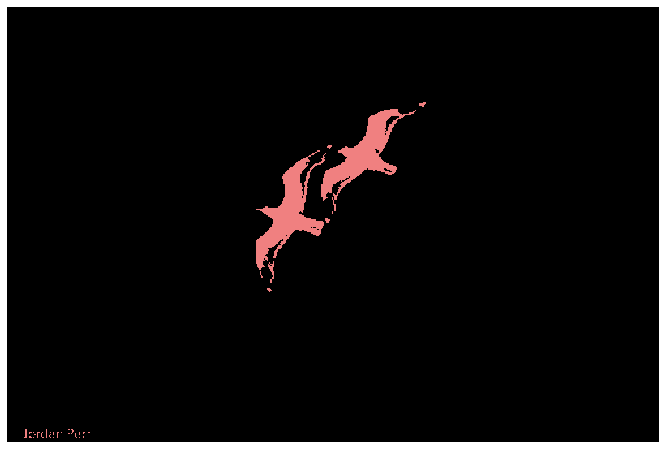} \\
		\caption*{mIOU=0.8892 }
	\end{minipage}
	\begin{minipage}{0.32\linewidth}
		\centering
		\caption*{Otsu}
		\includegraphics[height=0.08\textheight]{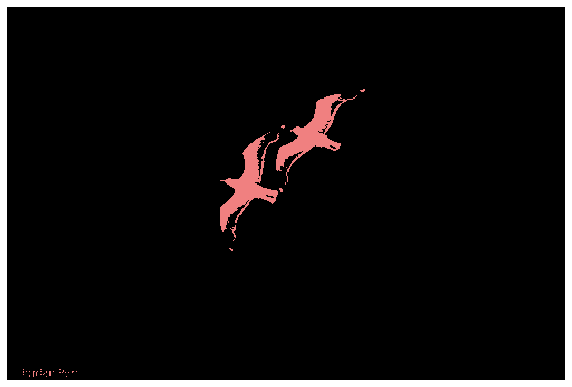} \\
		\caption*{mIOU=0.8663 }
	\end{minipage}
\begin{minipage}{0.32\linewidth}
	\centering
	\caption*{IQFT}
	\includegraphics[height=0.08\textheight]{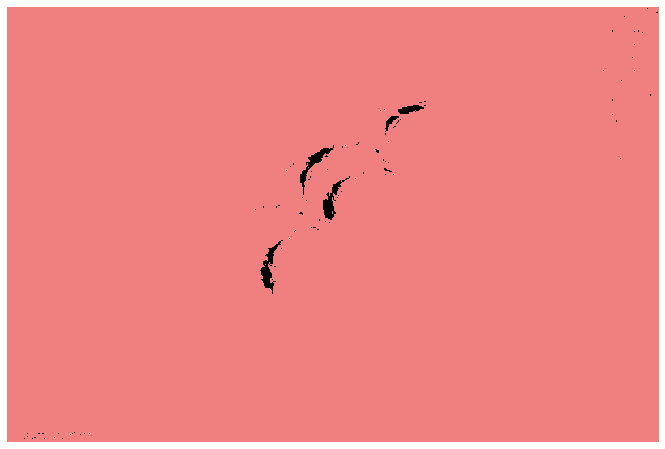} \\
	\caption*{mIOU=$\pi$,0.0084}
\end{minipage}
\begin{minipage}{0.32\linewidth}
	\centering
	\caption*{IQFT}
	\includegraphics[height=0.08\textheight]{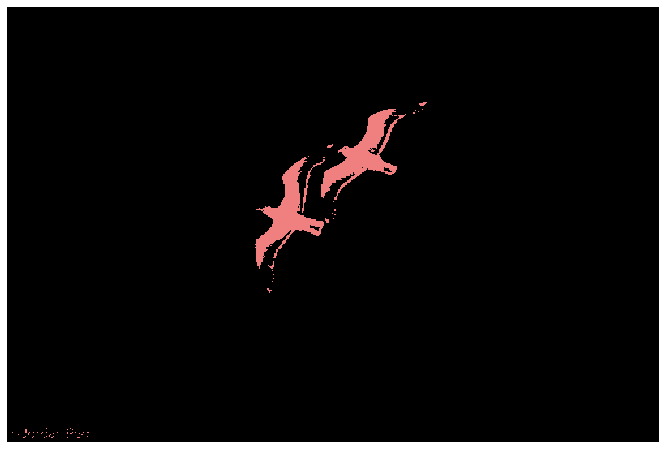} \\
	\caption*{mIOU=$3\pi/4$,0.8327}
\end{minipage}
	\caption{Performance improvement through $\theta$ adjustment. Using $\theta=3\pi/4$, rather than  $\theta=\pi$, can improve the segmentation quality of these images, as shown by increase in  the mIOU score }
	\label{fig:perform_improve2}
\end{figure}

The effectiveness of the proposed algorithm for image segmentation was validated by performing foreground-background segmentation on PASCAL VOC2012 and xVIEW2 challenge datasets. The resulting performance values are compared with K-means and Otsu-thresholding in Table~\ref{tab:PerfComp}. Based on the average mIOU values, the IQFT-inspired algorithm for RGB image segmentation was observed to outperform K-means and Otsu-thresholding in, respectively, 53.24\% and 52.32\% of the images in PASCAL VOC 2012. Similarly, the IQFT-inspired algorithm outperformed K-means and Otsu-thresholding in 95.94\% and 97.97\% of the pre-disaster images for  "joplin-tornado'' in the xVIEW2 challenge dataset. Figure~\ref{fig:perform_improve1} shows some segmentation outputs of PASCAL VOC 2012 images for which the IQFT-inspired algorithm outperformed the baseline techniques. Similar outputs are shown for xVIEW2 challenge dataset in Figure~\ref{fig:perform_xview}. However, the IQFT-inspired algorithm also showed poor performance($mIOU < 0.1$) for about 1.4\% of the PASCAL VOC 2012 images. This value doubles the observed values for K-means and Otsu thresholding because the performance of the proposed algorithm depends on the chosen value of angle parameter $\theta$  which was set to $ \pi $  in this experiment. Adjusting the value of $\theta$ for each image will result in  an observable performance improvement as shown in Figure~\ref{fig:perform_improve2}.

\begin{table*}[h]
	\centering
	\caption{Comparing the mIOU, computation time, and computational complexity. }
	\begin{tabular}{|l|l|c|c|c|c|}
		\hline
		~&~&	\multicolumn{4}{|c|}{Image segmentation methods} \\\cline{3-6}
		Datasets & Metrics & K-means & OTSU & IQFT& IQFT \\
		~ & ~ & ~ & ~ & (RGB) &(Grayscale)\\
		\hline \hline
		Pascal  & Average mIOU & 0.4318 & 0.4331 & \bfseries{0.4354} & \bfseries{0.4172}\\ \cline{2-6}
		VOC 2012    & Runtime (sec.)   & 0.25     & 0.01     & \bfseries{3.06} & \bfseries{1.76} \\ 
		\hline
		xVIEW2                 & Average mIOU   & 0.3375 & 0.4008 & \bfseries{0.5070}& \bfseries{0.478}  \\ \cline{2-6}
		~                     & Runtime (sec.)   & 1.74     & 0.10     & \bfseries{17.5}& \bfseries{9.67}  \\ 
		\hline 
	\end{tabular}
	\label{tab:PerfComp}
\end{table*}

\section{Conclusions}\label{sec:concl}
 In this work, a novel method for unsupervised image segmentation  based on the inverse quantum Fourier transform (IQFT) is proposed. 
 Specifically, the proposed method takes advantage of the phase information of the pixels in the image by encoding the pixels' intensity into qubit relative phases and applying IQFT to classify the pixels into different segments automatically and efficiently. 
 To the best of our knowledge, this is the first attempt of using IQFT for unsupervised image segmentation.
The proposed method has low computational cost comparing to the deep learning based methods and more importantly it does \emph{not} require training, thus make it suitable for real-time applications.
 Supported by the segmentation patterns obtained for image samples from xView2 and Pascal VOC datasets, this quantum-inspired method shows a promising performance when compared to the classical K-means clustering  and Otsu-thresholding methods. One of the appeals of the proposed method is that it automatically adapts to the characteristics of the image such  that the number of segments is not a required parameter like in  K-means. 
 
Note that the proposed method can be implemented in both classical computing domain and the quantum computing domain. Considering most of the datasets of image segmentation benchmark is in the classical computing domain, we have performed experiments to validate the proposed approach in the classical computing domain as well. We are working on the quantum domain implementation and the results will be shared in a future paper. 

\section{Acknowledgment}\label{sec:acknowledgement}
This research work is supported by the IBM-HBCU Quantum Center.


\begin{thebibliography}{10}

\bibitem{Color_image_segmentation_cheng2001}
H.-D. Cheng, X.~H. Jiang, Y.~Sun, and J.~Wang, ``Color image segmentation:
  advances and prospects,'' {\em Pattern recognition}, vol.~34, no.~12,
  pp.~2259--2281, 2001.

\bibitem{A_comparative_evaluation_for_liver_segmentation_from_spir_images_Goceri2013}
E.~Goceri, {\em A comparative evaluation for liver segmentation from spir
  images and a novel level set method using signed pressure force function}.
\newblock Izmir Institute of Technology (Turkey), 2013.

\bibitem{Image_segmentation_evaluation_wang2020}
Z.~Wang, E.~Wang, and Y.~Zhu, ``Image segmentation evaluation: a survey of
  methods,'' {\em Artificial Intelligence Review}, vol.~53, no.~8,
  pp.~5637--5674, 2020.

\bibitem{Segmentation_Techniques_for_Complex_Image_pandey2020}
R.~Pandey and R.~Lalchhanhima, ``Segmentation techniques for complex image,''
  in {\em 2020 International Conference on Computational Performance Evaluation
  (ComPE)}, pp.~804--808, IEEE, 2020.

\bibitem{Image_semantic_segmentation_method_based_on_improved_ERFNet_model_ye2022}
D.~Ye and R.~Han, ``Image semantic segmentation method based on improved erfnet
  model,'' {\em The Journal of Engineering}, vol.~2022, no.~2, pp.~180--190,
  2022.

\bibitem{A_survey_current_methods_in_medical_image_segmentation_Pham2000}
D.~L. Pham, C.~Xu, and J.~L. Prince, ``A survey of current methods in medical
  image segmentation,'' {\em Annual review of biomedical engineering}, vol.~2,
  no.~3, pp.~315--337, 2000.

\bibitem{Iteratively_learning_a_liver_segmentation_using_probabilistic_atlases_domingo2016}
J.~Domingo, E.~Dura, and E.~G{\"o}{\c{c}}eri, ``Iteratively learning a liver
  segmentation using probabilistic atlases: preliminary results,'' in {\em 2016
  15th IEEE international conference on machine learning and applications
  (ICMLA)}, pp.~593--598, IEEE, 2016.

\bibitem{High_resolution_encoder_decoder_networks_for_low_contrast_medical_image_segmentation_Zhou2019}
S.~Zhou, D.~Nie, E.~Adeli, J.~Yin, J.~Lian, and D.~Shen, ``High-resolution
  encoder--decoder networks for low-contrast medical image segmentation,'' {\em
  IEEE Transactions on Image Processing}, vol.~29, pp.~461--475, 2019.

\bibitem{Automated_fluorescent_miscroscopic_image_analysis_of_PTBP1_expression_in_glioma_kaya2017}
B.~Kaya, E.~Goceri, A.~Becker, B.~Elder, V.~Puduvalli, J.~Winter, M.~Gurcan,
  and J.~J. Otero, ``Automated fluorescent miscroscopic image analysis of ptbp1
  expression in glioma,'' {\em PLoS One}, vol.~12, no.~3, p.~e0170991, 2017.

\bibitem{Color_space_transformation_and_multiclass_weighted_loss_for_adhesive_white_blood_cell_segmentation_li2020}
H.~Li, X.~Zhao, A.~Su, H.~Zhang, J.~Liu, and G.~Gu, ``Color space
  transformation and multi-class weighted loss for adhesive white blood cell
  segmentation,'' {\em IEEE Access}, vol.~8, pp.~24808--24818, 2020.

\bibitem{Vessel_segmentation_from_abdominal_magnetic_resonance_images_goceri2017}
E.~Goceri, Z.~K. Shah, and M.~N. Gurcan, ``Vessel segmentation from abdominal
  magnetic resonance images: adaptive and reconstructive approach,'' {\em
  International journal for numerical methods in biomedical engineering},
  vol.~33, no.~4, p.~e2811, 2017.

\bibitem{Biomedical_information_technology_image_based_computer_aided_diagnosis_systems_goceri2018}
E.~Goceri and C.~Songul, ``Biomedical information technology: image based
  computer aided diagnosis systems,'' in {\em International Conference on
  Advanced Technologies, Antalaya, Turkey}, 2018.

\bibitem{Deep_Learning_approaches_biomedical_image_segmentation_Haque2020}
I.~R.~I. Haque and J.~Neubert, ``Deep learning approaches to biomedical image
  segmentation,'' {\em Informatics in Medicine Unlocked}, vol.~18, p.~100297,
  2020.

\bibitem{Hybrid_remote_sensing_image_segmentation_considering_intrasegment_homogeneity_and_intersegment_heterogeneity_wang2019}
Y.~Wang, Q.~Qi, L.~Jiang, and Y.~Liu, ``Hybrid remote sensing image
  segmentation considering intrasegment homogeneity and intersegment
  heterogeneity,'' {\em IEEE Geoscience and Remote Sensing Letters}, vol.~17,
  no.~1, pp.~22--26, 2019.

\bibitem{Hierarchical_weakly_supervised_learning_for_residential_area_semantic_segmentation_in_remote_sensing_images_zhang2019}
L.~Zhang, J.~Ma, X.~Lv, and D.~Chen, ``Hierarchical weakly supervised learning
  for residential area semantic segmentation in remote sensing images,'' {\em
  IEEE Geoscience and Remote Sensing Letters}, vol.~17, no.~1, pp.~117--121,
  2019.

\bibitem{Densely_based_multiscale_and_multimodal_fully_convolutional_networks_for_highresolution_remote_sensing_image_semantic_segmentation_peng2019}
C.~Peng, Y.~Li, L.~Jiao, Y.~Chen, and R.~Shang, ``Densely based multi-scale and
  multi-modal fully convolutional networks for high-resolution remote-sensing
  image semantic segmentation,'' {\em IEEE Journal of Selected Topics in
  Applied Earth Observations and Remote Sensing}, vol.~12, no.~8,
  pp.~2612--2626, 2019.

\bibitem{Aircraft_segmentation_from_remote_sensing_image_by_transferring_natual_image_trained_forground_extraction_CNN_model_zeng2019}
Y.~Zeng, X.~Niu, and Y.~Dou, ``Aircraft segmentation from remote sensing image
  by transferring natual image trained forground extraction cnn model,'' in
  {\em 2019 IEEE 4th international conference on signal and image processing
  (ICSIP)}, pp.~817--822, IEEE, 2019.

\bibitem{Dynamic_multicontext_segmentation_of_remote_sensing_images_based_on_convolutional_networks_nogueira2019}
K.~Nogueira, M.~Dalla~Mura, J.~Chanussot, W.~R. Schwartz, and J.~A. Dos~Santos,
  ``Dynamic multicontext segmentation of remote sensing images based on
  convolutional networks,'' {\em IEEE Transactions on Geoscience and Remote
  Sensing}, vol.~57, no.~10, pp.~7503--7520, 2019.

\bibitem{Optimal_segmentation_scale_selection_for_objectbased_change_detection_in_remote_sensing_images_using_Kullback_Leibler_divergencewu2019}
J.~Wu, B.~Li, W.~Ni, W.~Yan, and H.~Zhang, ``Optimal segmentation scale
  selection for object-based change detection in remote sensing images using
  kullback--leibler divergence,'' {\em IEEE Geoscience and Remote Sensing
  Letters}, vol.~17, no.~7, pp.~1124--1128, 2019.

\bibitem{Image_Decomposition_Accelerates_Dynamic_Network_Modeling_for_in_situ_Monitoring_of_Bio_mimic_Wing_Printing_Processes_aworunse2019}
O.~Aworunse, H.~Zhou, J.~Deng, and C.~Cheng, ``Image decomposition accelerates
  dynamic network modeling for in situ monitoring of bio-mimic wing printing
  processes,'' {\em Procedia Manufacturing}, vol.~39, pp.~178--184, 2019.

\bibitem{Texture_identification_and_image_segmentation_via_Fourier_transform_Zou2001}
M.~Zou and D.~Wang, ``Texture identification and image segmentation via fourier
  transform,'' in {\em Image Extraction, Segmentation, and Recognition},
  vol.~4550, pp.~34--39, SPIE, 2001.

\bibitem{Fundus_image_segmentation_and_feature_extraction_for_the_detection_of_glaucoma_fatima2018}
S.~T. Fatima~Bokhari, M.~Sharif, M.~Yasmin, and S.~L. Fernandes, ``Fundus image
  segmentation and feature extraction for the detection of glaucoma: A new
  approach,'' {\em Current Medical Imaging}, vol.~14, no.~1, pp.~77--87, 2018.

\bibitem{Joint_segmentation_and_nonlinear_registration_using_fft_and_total_variation_atta2018}
T.~Atta-Fosu and W.~Guo, ``Joint segmentation and nonlinear registration using
  fast fourier transform and total variation,'' in {\em Research in Shape
  Analysis}, pp.~111--132, Springer, 2018.

\bibitem{Coffee_plantation_area_recognition_in_satellite_images_using_Fourier_transformtsai2017}
D.-M. Tsai and W.-L. Chen, ``Coffee plantation area recognition in satellite
  images using fourier transform,'' {\em Computers and electronics in
  agriculture}, vol.~135, pp.~115--127, 2017.

\bibitem{Research_issues_on_digital_image_processing_for_various_applications_in_this_world_kannadhasan2014}
S.~Kannadhasan and V.~Bhapith, ``Research issues on digital image processing
  for various applications in this worldd,'' {\em Glob J Adv Res}, vol.~1,
  no.~1, pp.~46--55, 2014.

\bibitem{Challenges_and_recent_solutions_for_image_segmentation_in_the_era_of_deep_learning_goceri2019}
E.~Goceri, ``Challenges and recent solutions for image segmentation in the era
  of deep learning,'' in {\em 2019 ninth international conference on image
  processing theory, tools and applications (IPTA)}, pp.~1--6, IEEE, 2019.

\bibitem{Cell_segmentation_for_image_cytometry_advances_insufficiencies_and_challenges_wang2019}
Z.~Wang, ``Cell segmentation for image cytometry advances insufficiencies and
  challenges,'' {\em Cytometry A}, vol.~95, no.~7, pp.~708--711, 2019.

\bibitem{Image_segmentation_available_techniques_open_issues_and_region_growing_algorithm_ghule2012}
A.~Ghule and P.~Deshmukh, ``Image segmentation available techniques open issues
  and region growing algorithm,'' {\em Journal of Signal and Image Processing},
  vol.~3, no.~1, 2012.

\bibitem{A_comparative_performance_study_of_several_global_thresholding_techniques_for_segmentation_lee1990}
S.~U. Lee, S.~Y. Chung, and R.~H. Park, ``A comparative performance study of
  several global thresholding techniques for segmentation,'' {\em Computer
  Vision, Graphics, and Image Processing}, vol.~52, no.~2, pp.~171--190, 1990.

\bibitem{Image_segmentation_by_three_level_thresholding_based_on_maximum_fuzzy_entropy_and_genetic_algorithm_tao2003}
W.-B. Tao, J.-W. Tian, and J.~Liu, ``Image segmentation by three-level
  thresholding based on maximum fuzzy entropy and genetic algorithm,'' {\em
  Pattern Recognition Letters}, vol.~24, no.~16, pp.~3069--3078, 2003.

\bibitem{Region_based_segmentation_versus_edge_detection_kaganami2009}
H.~G. Kaganami and Z.~Beiji, ``Region-based segmentation versus edge
  detection,'' in {\em 2009 Fifth International Conference on Intelligent
  Information Hiding and Multimedia Signal Processing}, pp.~1217--1221, IEEE,
  2009.

\bibitem{Application_of_region_based_segmentation_and_neural_network_edge_detection_to_skin_lesionsrajab2004}
M.~Rajab, M.~Woolfson, and S.~Morgan, ``Application of region-based
  segmentation and neural network edge detection to skin lesions,'' {\em
  Computerized Medical Imaging and Graphics}, vol.~28, no.~1-2, pp.~61--68,
  2004.

\bibitem{Edge_region_based_segmentation_of_range_images_wani1994}
M.~A. Wani and B.~G. Batchelor, ``Edge-region-based segmentation of range
  images,'' {\em IEEE Transactions on Pattern Analysis and Machine
  Intelligence}, vol.~16, no.~3, pp.~314--319, 1994.

\bibitem{Edge_and_region_based_segmentation_technique_for_the_extraction_of_large_2004}
M.~Mueller, K.~Segl, and H.~Kaufmann, ``Edge-and region-based segmentation
  technique for the extraction of large, man-made objects in high-resolution
  satellite imagery,'' {\em Pattern recognition}, vol.~37, no.~8,
  pp.~1619--1628, 2004.

\bibitem{Survey_on_clustering_based_image_segmentation_techniques_zou2016}
Y.~Zou and B.~Liu, ``Survey on clustering-based image segmentation
  techniques,'' in {\em 2016 IEEE 20th International Conference on Computer
  Supported Cooperative Work in Design (CSCWD)}, pp.~106--110, IEEE, 2016.

\bibitem{Analysis_of_color_images_using_cluster_based_segmentation_techniques_mohanty2013}
A.~Mohanty, S.~Rajkumar, Z.~M. Mir, and P.~Bardhan, ``Analysis of color images
  using cluster based segmentation techniques,'' {\em International Journal of
  Computer Applications}, vol.~79, no.~2, 2013.

\bibitem{Comparative_study_of_clustering_based_colour_image_segmentation_techniques_chebbout2012}
S.~Chebbout and H.~F. Merouani, ``Comparative study of clustering based colour
  image segmentation techniques,'' in {\em 2012 Eighth International Conference
  on Signal Image Technology and Internet Based Systems}, pp.~839--844, IEEE,
  2012.

\bibitem{Image_segmentation_based_on_watershed_and_edge_detection_techniques_salman2006}
N.~Salman, ``Image segmentation based on watershed and edge detection
  techniques,'' {\em Int. Arab J. Inf. Technol.}, vol.~3, no.~2, pp.~104--110,
  2006.

\bibitem{Generalized_multi_task_learning_from_substantially_unlabeled_multi_source_medical_image_data_haque2021}
A.~Haque, A.-A.-Z. Imran, A.~Wang, and D.~Terzopoulos, ``Generalized multi-task
  learning from substantially unlabeled multi-source medical image data,'' {\em
  arXiv preprint arXiv:2110.13185}, 2021.

\bibitem{A_review_on_deep_learning_techniques_applied_to_semantic_segmentation_garcia2017}
A.~Garcia-Garcia, S.~Orts-Escolano, S.~Oprea, V.~Villena-Martinez, and
  J.~Garcia-Rodriguez, ``A review on deep learning techniques applied to
  semantic segmentation,'' {\em arXiv preprint arXiv:1704.06857}, 2017.

\bibitem{Understanding_deep_learning_techniques_for_image_segmentation_ghosh2019}
S.~Ghosh, N.~Das, I.~Das, and U.~Maulik, ``Understanding deep learning
  techniques for image segmentation,'' {\em ACM Computing Surveys (CSUR)},
  vol.~52, no.~4, pp.~1--35, 2019.

\bibitem{Survey_of_recent_progress_in_semantic_image_segmentation_with_CNNs_geng2018}
Q.~Geng, Z.~Zhou, and X.~Cao, ``Survey of recent progress in semantic image
  segmentation with cnns,'' {\em Science China Information Sciences}, vol.~61,
  no.~5, pp.~1--18, 2018.

\bibitem{A_threshold_selection_method_from_graylevel_histograms_otsu1979}
N.~Otsu, ``A threshold selection method from gray-level histograms,'' {\em IEEE
  transactions on systems, man, and cybernetics}, vol.~9, no.~1, pp.~62--66,
  1979.

\bibitem{Classification_and_analysis_of_multivariate_observations_macqueen1967}
J.~MacQueen, ``Classification and analysis of multivariate observations,'' in
  {\em 5th Berkeley Symp. Math. Statist. Probability}, pp.~281--297, 1967.

\bibitem{The_dual_threshold_quantum_image_segmentation_algorithm_and_its_simulation_yuan2020}
S.~Yuan, C.~Wen, B.~Hang, and Y.~Gong, ``The dual-threshold quantum image
  segmentation algorithm and its simulation,'' {\em Quantum Information
  Processing}, vol.~19, no.~12, pp.~1--21, 2020.

\bibitem{Multilevel_image_threshold_segmentation_using_an_improved_Bloch_quantum_artificial_bee_colony_algorithmhuo2020}
F.~Huo, X.~Sun, and W.~Ren, ``Multilevel image threshold segmentation using an
  improved bloch quantum artificial bee colony algorithm,'' {\em Multimedia
  Tools and Applications}, vol.~79, no.~3, pp.~2447--2471, 2020.

\bibitem{Design_of_threshold_segmentation_method_for_quantum_image_Li2020}
P.~Li, T.~Shi, Y.~Zhao, and A.~Lu, ``Design of threshold segmentation method
  for quantum image,'' {\em International Journal of Theoretical Physics},
  vol.~59, no.~2, pp.~514--538, 2020.

\bibitem{Image_segmentation_on_a_quantum_computer_caraiman2015}
S.~Caraiman and V.~I. Manta, ``Image segmentation on a quantum computer,'' {\em
  Quantum Information Processing}, vol.~14, no.~5, pp.~1693--1715, 2015.

\bibitem{Image_storage_retrieval_compression_and_segmentation_in_a_quantum_system_li2013}
H.-S. Li, Z.~Qingxin, S.~Lan, C.-Y. Shen, R.~Zhou, and J.~Mo, ``Image storage,
  retrieval, compression and segmentation in a quantum system,'' {\em Quantum
  information processing}, vol.~12, no.~6, pp.~2269--2290, 2013.

\bibitem{Optimal_quantum_phase_estimation_dorner2009}
U.~Dorner, R.~Demkowicz-Dobrzanski, B.~J. Smith, J.~S. Lundeen, W.~Wasilewski,
  K.~Banaszek, and I.~A. Walmsley, ``Optimal quantum phase estimation,'' {\em
  Physical review letters}, vol.~102, no.~4, p.~040403, 2009.

\bibitem{Quantum_mechanics_helps_in_searching_for_a_needle_in_a_haystack_grover1997}
L.~K. Grover, ``Quantum mechanics helps in searching for a needle in a
  haystack,'' {\em Physical review letters}, vol.~79, no.~2, p.~325, 1997.

\bibitem{Algorithms_for_quantum_computation_discrete_logarithms_and_factoring_shor1994}
P.~W. Shor, ``Algorithms for quantum computation: discrete logarithms and
  factoring,'' in {\em Proceedings 35th annual symposium on foundations of
  computer science}, pp.~124--134, Ieee, 1994.

\bibitem{Mixed_state_entanglement_and_quantum_error_correction_bennett1996}
C.~H. Bennett, D.~P. DiVincenzo, J.~A. Smolin, and W.~K. Wootters,
  ``Mixed-state entanglement and quantum error correction,'' {\em Physical
  Review A}, vol.~54, no.~5, p.~3824, 1996.

\bibitem{System_and_method_for_key_distribution_using_quantum_cryptography_townsend1997}
P.~D. Townsend, ``System and method for key distribution using quantum
  cryptography,'' Oct.~7 1997.
\newblock US Patent 5,675,648.

\bibitem{Insights_into_the_Viability_of_Using_Available_Photonic_Quantum_Technologies_for_Efficient_iliyasu2013}
A.~M. Iliyasu, P.~Q. Le, F.~Yan, B.~Sun, F.~Dong, A.~K. Al-Asmari, and
  K.~Hirota, ``Insights into the viability of using available photonic quantum
  technologies for efficient image and video processing applications.,'' {\em
  International Journal of Unconventional Computing}, vol.~9, 2013.

\bibitem{Implementation_and_Analysis_of_QFT_in_Image_Processing_al}
O.~Al-Ta’ani1a, A.~M. Alqudah, and M.~Al-Bzoor, ``Implementation and analysis
  of quantum fourier transform in image processing,''

\bibitem{Quantum_Computation_Quantum_Information_Nielsen_Chuang_2011}
M.~A. Nielsen and I.~L. Chuang, {\em Quantum Computation and Quantum
  Information: 10th Anniversary Edition}.
\newblock Cambridge University Press, 2011.

\bibitem{Multi_dimensional_quantum_state_sharing_based_on_QFT_qin2018}
H.~Qin, R.~Tso, and Y.~Dai, ``Multi-dimensional quantum state sharing based on
  quantum fourier transform,'' {\em Quantum Information Processing}, vol.~17,
  no.~3, pp.~1--12, 2018.

\bibitem{Quantum_arithmetic_with_QFT_ruiz2017}
L.~Ruiz-Perez and J.~C. Garcia-Escartin, ``Quantum arithmetic with the quantum
  fourier transform,'' {\em Quantum Information Processing}, vol.~16, no.~6,
  pp.~1--14, 2017.

\bibitem{Unsupervised_learning_of_image_segmentation_based_on_differentiable_feature_clustering_kim2020}
W.~Kim, A.~Kanezaki, and M.~Tanaka, ``Unsupervised learning of image
  segmentation based on differentiable feature clustering,'' {\em IEEE
  Transactions on Image Processing}, vol.~29, pp.~8055--8068, 2020.

\bibitem{xbd_A_dataset_for_assessing_building_damage_from_satellite_imagery_gupta2019}
R.~Gupta, R.~Hosfelt, S.~Sajeev, N.~Patel, B.~Goodman, J.~Doshi, E.~Heim,
  H.~Choset, and M.~Gaston, ``xbd: A dataset for assessing building damage from
  satellite imagery,'' {\em arXiv preprint arXiv:1911.09296}, 2019.

\bibitem{scikit_image_mage_processing_in_Python_van_2014}
S.~Van~der Walt, J.~L. Sch{\"o}nberger, J.~Nunez-Iglesias, F.~Boulogne, J.~D.
  Warner, N.~Yager, E.~Gouillart, and T.~Yu, ``scikit-image: image processing
  in python,'' {\em PeerJ}, vol.~2, p.~e453, 2014.

\bibitem{Scikit_learn_Machine_Learning_in_Python_Pedregosa_2011}
F.~Pedregosa, G.~Varoquaux, A.~Gramfort, V.~Michel, B.~Thirion, O.~Grisel,
  M.~Blondel, P.~Prettenhofer, R.~Weiss, V.~Dubourg, J.~Vanderplas, A.~Passos,
  D.~Cournapeau, M.~Brucher, M.~Perrot, and E.~Duchesnay, ``Scikit-learn:
  Machine learning in python,'' {\em Journal of Machine Learning Research},
  vol.~12, pp.~2825--2830, 2011.

\bibitem{LargeScale_Machine_Learning_on_Heterogeneous_Systems_tensorflow2015}
M.~Abadi, A.~Agarwal, P.~Barham, E.~Brevdo, Z.~Chen, C.~Citro, G.~S. Corrado,
  A.~Davis, J.~Dean, M.~Devin, S.~Ghemawat, I.~Goodfellow, A.~Harp, G.~Irving,
  M.~Isard, Y.~Jia, R.~Jozefowicz, L.~Kaiser, M.~Kudlur, J.~Levenberg,
  D.~Man\'{e}, R.~Monga, S.~Moore, D.~Murray, C.~Olah, M.~Schuster, J.~Shlens,
  B.~Steiner, I.~Sutskever, K.~Talwar, P.~Tucker, V.~Vanhoucke, V.~Vasudevan,
  F.~Vi\'{e}gas, O.~Vinyals, P.~Warden, M.~Wattenberg, M.~Wicke, Y.~Yu, and
  X.~Zheng, ``{TensorFlow}: Large-scale machine learning on heterogeneous
  systems,'' 2015.
\newblock Software available from tensorflow.org.

\bibitem{The_pascal_visual_object_classes_challenge_A_retrospective_everingham2015}
M.~Everingham, S.~Eslami, L.~Van~Gool, C.~K. Williams, J.~Winn, and
  A.~Zisserman, ``The pascal visual object classes challenge: A
  retrospective,'' {\em International journal of computer vision}, vol.~111,
  no.~1, pp.~98--136, 2015.

\end{thebibliography}

\end{document}